\documentclass[runningheads]{llncs}
\usepackage{graphicx}

\usepackage{tikz}
\usepackage{comment}
\usepackage{amsmath,amssymb} 
\usepackage{color}

\usepackage{multirow}

\usepackage{wrapfig}

\usepackage{caption} 
\usepackage[accsupp]{axessibility}  

\usepackage[width=122mm,left=12mm,paperwidth=146mm,height=193mm,top=12mm,paperheight=217mm]{geometry}

\usepackage{kotex}
\usepackage{enumitem}
\definecolor{nicergreen}{rgb}{0.13, 0.54, 0.13}
\definecolor{nicered}{rgb}{0.83, 0.16, 0.16}

\newcommand{\fref}[1]{Fig. \ref{#1}}
\newcommand{\eref}[1]{Eq. (\ref{#1})}
\newcommand{\tref}[1]{Table \ref{#1}}
\newcommand{\sref}[1]{$\S$ \ref{#1}}
\newcommand{\aref}[1]{Appendix \ref{#1}}

\newcommand{\x}{\mathbf{x}}
\newcommand{\xfg}{\x_\text{fg}}
\newcommand{\xbg}{\x_\text{bg}}
\newcommand{\xcomp}{\x_\text{comp}}
\newcommand{\m}{\mathbf{m}}

\newcommand{\G}{\mathcal G}
\newcommand{\D}{\mathcal D}
\newcommand{\gfg}{\G_\text{fg}}
\newcommand{\gm}{\G_\text{mask}}
\newcommand{\gbg}{\G_\text{bg}}

\newcommand{\z}{\mathbf{z}}
\newcommand{\zfg}{\z_\text{fg}}
\newcommand{\zbg}{\z_\text{bg}}

\newcommand{\daux}{\D_\text{aux}}
\newcommand{\sg}{\texttt{stopgrad}}

\newcommand{\mc}{\m_\text{coarse}}
\newcommand{\mf}{\m_\text{fine}}
\newcommand{\hmf}{\tilde{\m}_\text{fine}}
\newcommand{\gmc}{\G_{\mc}}
\newcommand{\gmf}{\G_{\mf}}
\newcommand{\f}{\mathbf{f}}

\begin{document}
\pagestyle{headings}
\mainmatter
\def\ECCVSubNumber{408}  

\title{FurryGAN: High Quality Foreground-aware Image Synthesis}

\titlerunning{FurryGAN}
%
\author{Jeongmin Bae \and
Mingi Kwon \and
Youngjung Uh\thanks{Corresponding author}}
\authorrunning{J. Bae et al.}
%
\institute{Yonsei University\\
\email{\{jaymin.bae, kwonmingi, yj.uh\}@yonsei.ac.kr}}
\maketitle

\ifx\supp\undefined

\begin{figure}[h!]
\centering
\includegraphics[width=\linewidth]{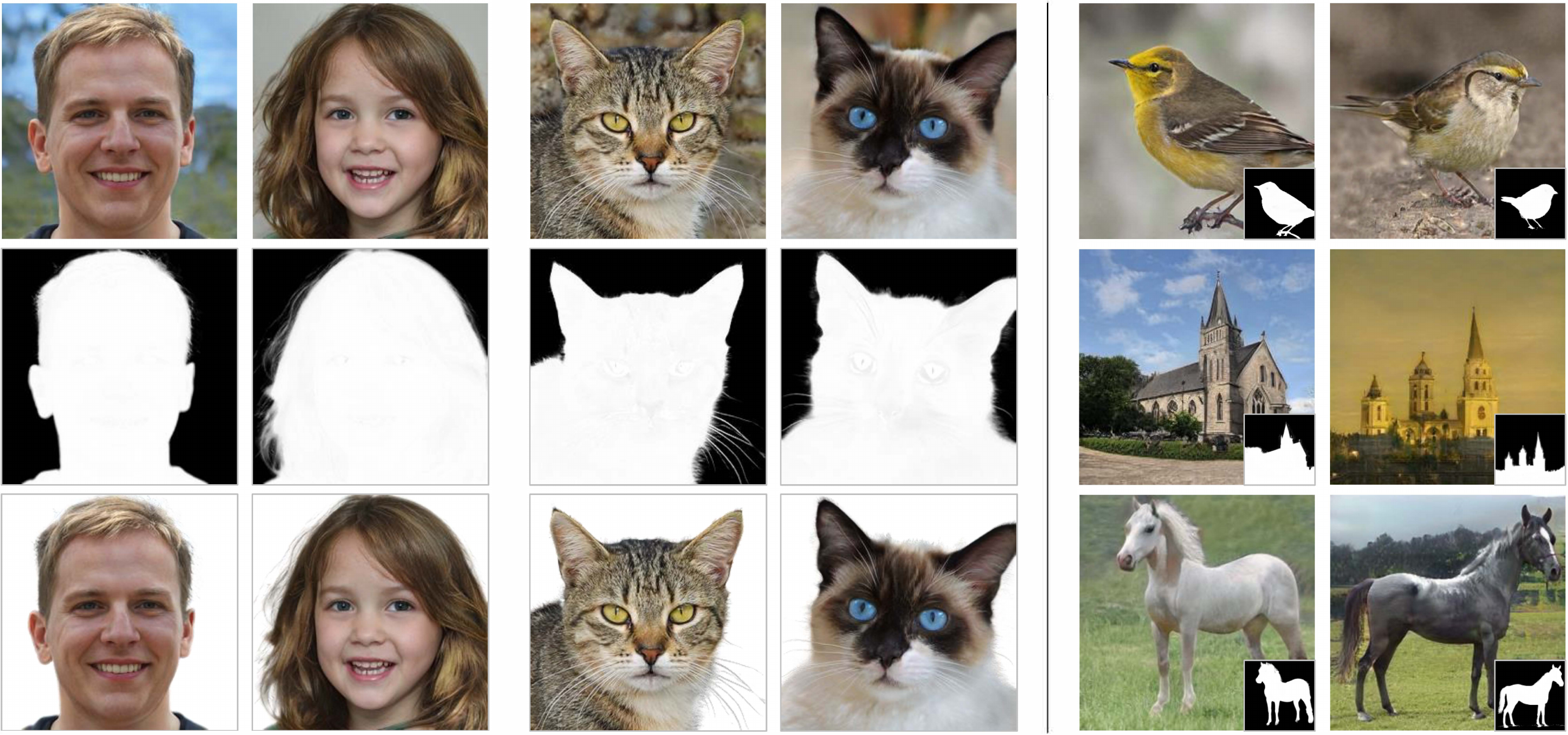}
\caption{\textbf{Example images and the corresponding foreground masks. Both are simultaneously \textit{generated} by our model.}
FurryGAN learns not only to generate realistic images, but also to synthesize alpha masks with fine details such as hair, fur, and whiskers in a fully unsupervised manner (left).
Our model also can be trained on various datasets (right).
}
\label{fig:teaser}
\end{figure}

\begin{abstract}
Foreground-aware image synthesis aims to generate images as well as their foreground masks. A common approach is to formulate an image as an masked blending of a foreground image and a background image.
It is a challenging problem because it is prone to reach the trivial solution where either image overwhelms the other, i.e., the masks become completely full or empty, and the foreground and background are not meaningfully separated.
We present FurryGAN with three key components: 1) imposing both the foreground image and the composite image to be realistic, 2) designing a mask as a combination of coarse and fine masks, and 3) guiding the generator by an auxiliary mask predictor in the discriminator.
Our method produces realistic images with remarkably detailed alpha masks which cover hair, fur, and whiskers in a fully unsupervised manner. 
Project page: \url{https://jeongminb.github.io/FurryGAN/}
\end{abstract}

\section{Introduction}
As the quality of images from generative adversarial networks (GANs) improves \cite{goodfellow2014generative,karras2019style,karras2020analyzing,karras2020training,karras2021alias}, discovering the semantics in their latent space is useful to control the generation process \cite{shen2020interfacegan,abdal2021styleflow,shen2021closed,yuksel2021latentclr} or to edit real images through latent inversion \cite{patashnik2021styleclip,richardson2021encoding,zhu2020domain,alaluf2021restyle}. 
Localizing the semantics in the latent space is another important research direction for understanding how GANs work. Some methods tackle local editing by separating parts in the intermediate feature maps \cite{collins2020editing,kim2021exploiting}. 

Meanwhile, a few recent works tackle foreground-aware image synthesis by modeling an image as a composition of foreground and background images according to a mask.
While previous methods achieve some success, they explicitly prepare a background distribution by removing foreground with an off-the-shelf object detector \cite{singh2019finegan}, assume that images with shifted foreground objects should look real \cite{bielski2019emergence,yang2022learning}, or require multi-stage training with dataset-tailored hyperparameters \cite{abdal2021labels4free}. These ingredients are obstacles that block general solutions for foreground-aware synthesis.

In this paper, we propose FurryGAN which learns to synthesize images with the explicit understanding of the foreground given only a collection of images. Intuitions in our method include the following. 1) We encourage the foreground images and the composite images to resemble the training distribution. It prevents the foreground from losing the objects. 2) We introduce coarse and fine masks. The coarse mask captures the rough shape, and the fine mask captures details such as whiskers and hair. 3) We introduce an auxiliary task for the discriminator to predict the mask from the generated image
so that the generator produces the foreground image aligned with the mask.

Compared to the previous works, our method does not require off-the-shelf networks, the assumption for perturbation, multi-stage training, or careful early stopping. 
Experiments demonstrate the superiority of our framework compared to previous methods regarding high quality alpha masks. We also provide thorough ablation studies to justify each component of our method.

\fref{fig:teaser} shows example synthesized images and the corresponding alpha masks. They catch unprecedented levels of fine details, especially in hair and whiskers. Consequently, the detailed masks enable natural composition of the foreground part and any background (\fref{fig:composite}). As a byproduct, GAN inversion on our method achieves unsupervised object segmentation with the same level of details.


\section{Related Work}
\subsubsection{GANs and semantic interpretation.}
GANs \cite{karras2019style,karras2020analyzing,karras2021alias} synthesize astonishingly high quality images from random latent codes. Understanding semantic interpretation of the latent codes is an important research topic so that users can control the generation process or edit real images through latent inversion \cite{shen2020interfacegan,zhu2020domain,tov2021designing,richardson2021encoding,wu2021stylespace}. Instead, we focus on teaching GANs spatial understanding of foreground objects.

\subsubsection{Foreground-aware GANs.}
Although semantic interpretation has some correlation with spatial separation, incorporating the notion of foreground objects has been tackled in the orthogonal direction, mostly by modeling an image as a combination of foreground and background according to a mask. PSeg \cite{bielski2019emergence} and improved layered GAN \cite{yang2022learning} rely on an assumption that the composite image with spatially transformed foreground should still be realistic. 
However, the parameters for the transformation should be determined for each dataset, and the assumption does not hold when the foreground region touches a border of the image. 
In \cite{voynov2020unsupervised,melas2021finding,voynov2021object}, they identify the latent directions in a pretrained generator for changing the background to separate the foreground and background. Labels4Free \cite{abdal2021labels4free} trains an alpha mask network that produces masks for combining foregrounds and backgrounds, generated by pretrained StyleGAN2 and pretrained pseudo-background StyleGAN2, respectively. 
Whereas it requires multi-stage training with tailored hyperparameters, our framework is trained in end-to-end fashion and produces remarkably fine details in the masks. 
\subsubsection{Unsupervised segmentation.}
Early image segmentation methods rely on the clustering of color and coordinates \cite{achanta2012slic,comaniciu2002mean}. In order to cluster the regions regarding semantics, learning deep networks for maximizing mutual information within the cluster \cite{ji2019invariant,ouali2020autoregressive} or for contrasting different instances \cite{van2021unsupervised} have been successful. These objectives assume multiple classes and are not straightforward to be applied in foreground-background separation. Given a generator for foreground-aware image synthesis, inverting a real image to the latent space inherently leads to unsupervised foreground segmentation. Thus we focus on a better understanding of foreground in GANs.

\subsubsection{3D-aware GANs.}
3D-aware GANs based on NeRF \cite{mildenhall2020nerf} represent a scene as a neural network which receives 3D coordinates and outputs their color or feature vector with occupancy. 
Furthermore, recent approaches divide the scene into foreground and background by a depth threshold \cite{gu2021stylenerf,zhang2020nerf++} or separate feature fields \cite{niemeyer2021giraffe}.
However, they aim to understand the 3D geometry of the scene and do not explicitly learn to generate high quality foreground alpha masks.


\begin{figure*}[!t]
\begin{center}
    \includegraphics[width=0.9\linewidth]{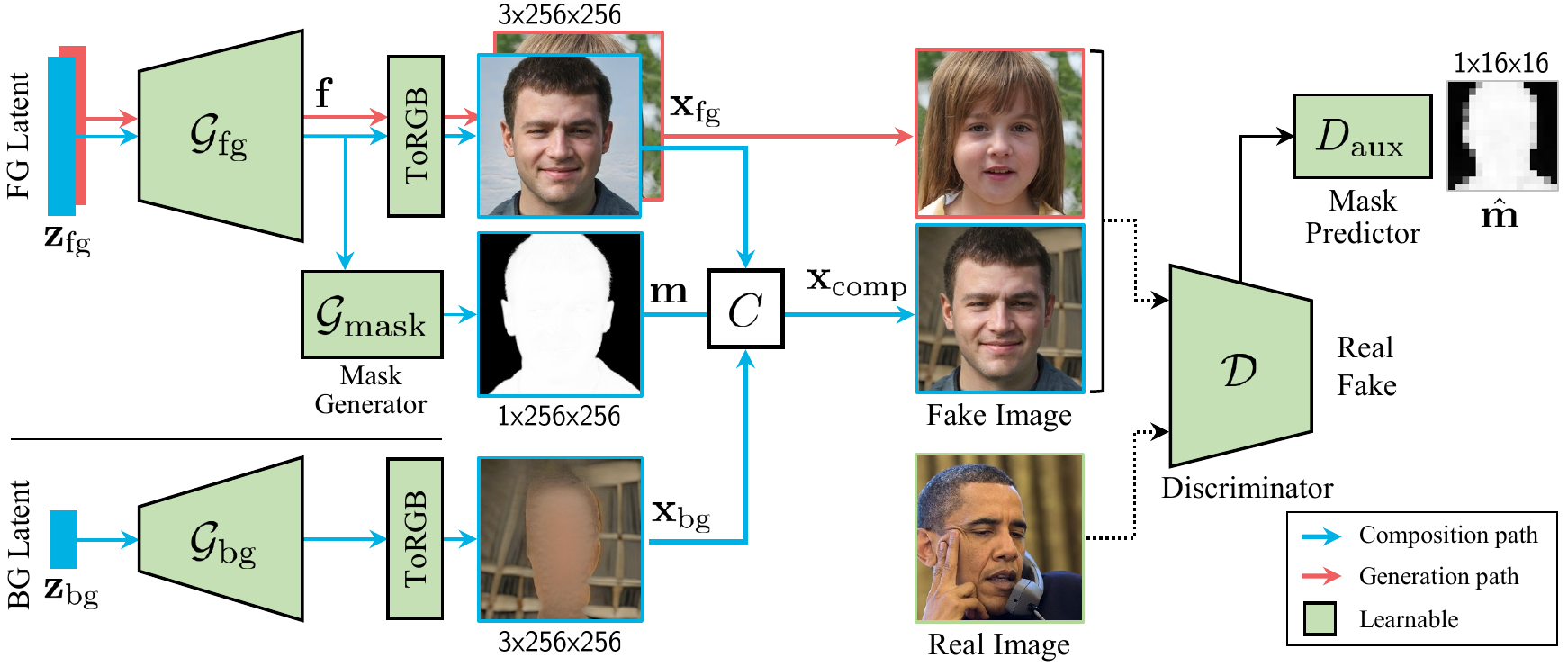}
    \caption{
    \textbf{Our framework.} consists of a foreground generator, a mask generator, a background generator, and a discriminator with a mask predictor. The alpha mask specifies the combination of the foreground image and the background image to produce the composite image. We feed both the foreground images and the composite images to the discriminator as fake images.
    }
    \label{fig:framework}
\end{center}
\end{figure*}

\section{Method}
In this section, we overview our framework (\sref{sec:framework}), describe the networks (\sref{sec:arch}), and explain their training techniques including loss functions (\sref{sec:training}). 
To begin with, we briefly introduce a common formulation.


\subsubsection{Common formulation.} \label{common formulation}
We follow the common formulation \cite{abdal2021labels4free,bielski2019emergence,singh2019finegan,yang2022learning} for generating images: an image is a masked combination of a foreground image $\xfg$ and a background image $\xbg$, according to an alpha mask $\m$. Formally,

\begin{equation} \label{eq:composite}
\xcomp = \m \odot \xfg  + (1 - \m) \odot \xbg,
\end{equation}
where $\odot$ denotes pixel-wise multiplication.

\subsection{Framework overview and dual fake input strategy} \label{sec:framework}

\fref{fig:framework} shows the framework overview. FurryGAN has three generators for the foreground, background, and mask to produce the composite images according to \eref{eq:composite}. Then the discriminator guides the generator to produce realistic images. 


\subsubsection{Dual fake input strategy.}

Guiding the generator to produce realistic \textit{composite} images solely does not guarantee the separation of the foreground and background. 
Motivation of the dual fake input is the following. The foreground images should contain salient objects (e.g., a person in FFHQ) so that there exists a solution for the masks to produce realistic composite images including the foreground images. Otherwise, the mask will favor excluding the foreground images from the composite images. Hence, we ensure the foreground images to contain salient objects by imposing a sufficient condition: being realistic by themselves. The fake mini-batch for the discriminator consists of the foreground images and the composite images (\fref{fig:method}(a)).
Then the discriminator tries to classify them as fakes and the generator tries to produce realistic images in both the foreground and the composite images. We find that the dual fake input strategy helps prevent improper foreground separation.

\begin{figure*}[t]
\begin{center}
    \includegraphics[width=0.9\linewidth]{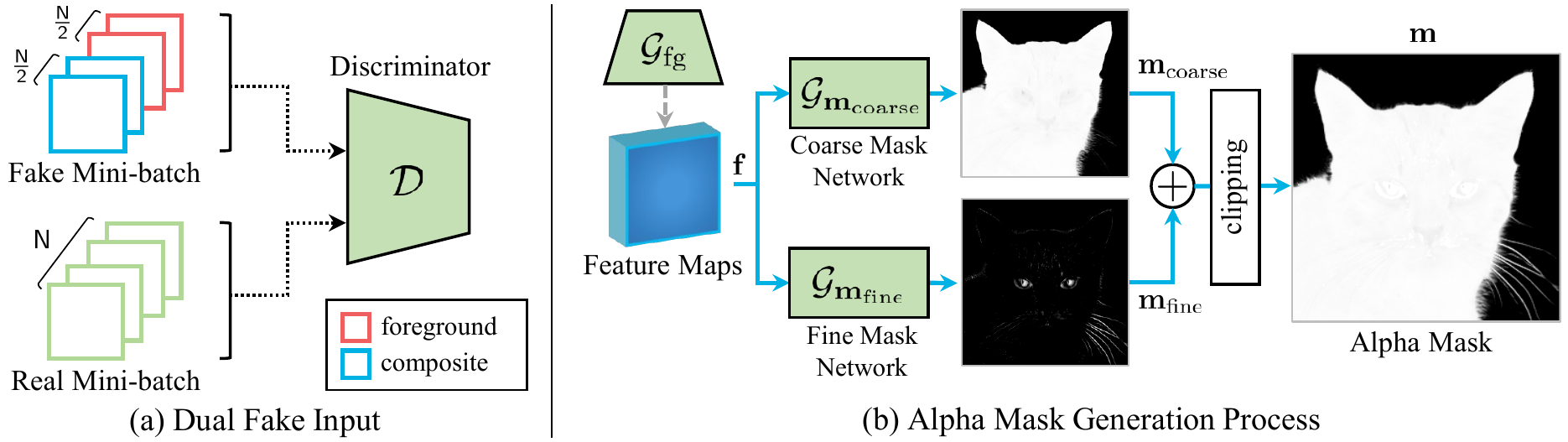}
    \caption{
    \textbf{Dual fake input strategy and mask generators.} (a) Our discriminator receives the foreground images and the composite images as fake. They evenly share a fake mini-batch. (b) Our mask generator produces an alpha mask as a combination of a coarse mask and a fine mask.}
    \label{fig:method}
\end{center}
\end{figure*}

\subsection{Architecture} \label{sec:arch}
\subsubsection{Generators.}
The foreground and background generators, $\gfg$ and $\gbg$, synthesize images $\xfg$ and $\xbg$ from latent codes $\zfg$ and $\zbg$, respectively. The two generators do not share any parameters. The mask generator $\gm$ synthesizes $\m$ from the penultimate feature maps of the foreground generator. Then, their simple alpha-blending produces the composite image $\xcomp$ (\eref{eq:composite}).
Note that the composite function causes unexpected additional degree of freedom: $\xfg\odot\m=2\cdot\xfg\odot0.5\cdot\m$. Thus we restrict the generators' outputs to be in range of $[-1,1]$ by adding a tanh function at the output of the ToRGB layer. 

\subsubsection{Coarse and fine mask generator.}
As shown in \fref{fig:method} (b), our mask generator consists of a coarse mask network $\gmc$ and a fine mask network $\gmf$. 
We expect the coarse mask network to cover the overall shape and the fine mask network to make up for the details 
missed in the coarse mask (e.g., cat whiskers, fur, and hair). Each mask is normalized to the range of [0,1] by min-max normalization and their summation becomes the final alpha mask. The final mask $\m$ is computed as:
\begin{equation}
\mc = \gmc(\f), \quad \mf = \gmf(\f),
\end{equation}
\begin{equation}
\label{eq:final_mask}
\m = \text{clip}(\mc + \gamma \mf, 0, 1),
\end{equation}
where $\f$ denotes the penultimate feature maps of the foreground generator.
The design details are described in the appendix (\sref{appendix:masks}).
For stability, we fade in the fine mask by linearly increasing $\gamma$ from 0 to 1 over the first 5K iterations.

\subsubsection{Discriminator with a mask predictor.} \label{sec:discriminator}
We follow the discriminator architecture of StyleGAN2 \cite{karras2020analyzing} and add an auxiliary mask predictor. The mask predictor tries to reconstruct the mask of an input image given the $16\times16$ feature maps. It has minimal capacity for predicting the masks, i.e., two $1\times1$ convolutional layers and residual connections.
How it guides the generators will be discussed in the following section (\eref{eq:mask_prediction} and \eref{eq:mask_consistency}).

\subsection{Training objectives} \label{sec:training}
\subsubsection{Adversarial loss.}
As described in \sref{sec:framework}, we impose adversarial losses  on the foreground image and the composite image. We adopt non-saturating loss \cite{goodfellow2014generative} and lazy R1-regularization \cite{mescheder2018training,karras2020analyzing} and skip defining trivial equations $L^\D_\text{adv}$, $L^\D_\text{R1}$, and $L^\G_\text{adv}$ for brevity. Adversarial losses act as the primary source for driving foreground-aware image synthesis. 

\subsubsection{Mask prediction loss.} The auxiliary mask predictor $\daux$ in the discriminator aims to regress the generated mask given the generated image:
\begin{equation} \label{eq:mask_prediction}
L_{\text{pred}}= \frac{1}{|\hat \m|} \| \texttt{Downsample}(\m) -  \hat{\m}\|^2_2,
\end{equation}
where $\hat{\m}$ is the output of $\daux$ for the generated images ($\xfg$ and $\xcomp$).
We use bilinear interpolation for \texttt{Downsample}. The 16$\times$16 prediction will be useful for guiding the generator in conjunction with \eref{eq:mask_consistency}.

\begin{figure*}[t]
\begin{center}
    \includegraphics[width=0.9\linewidth]{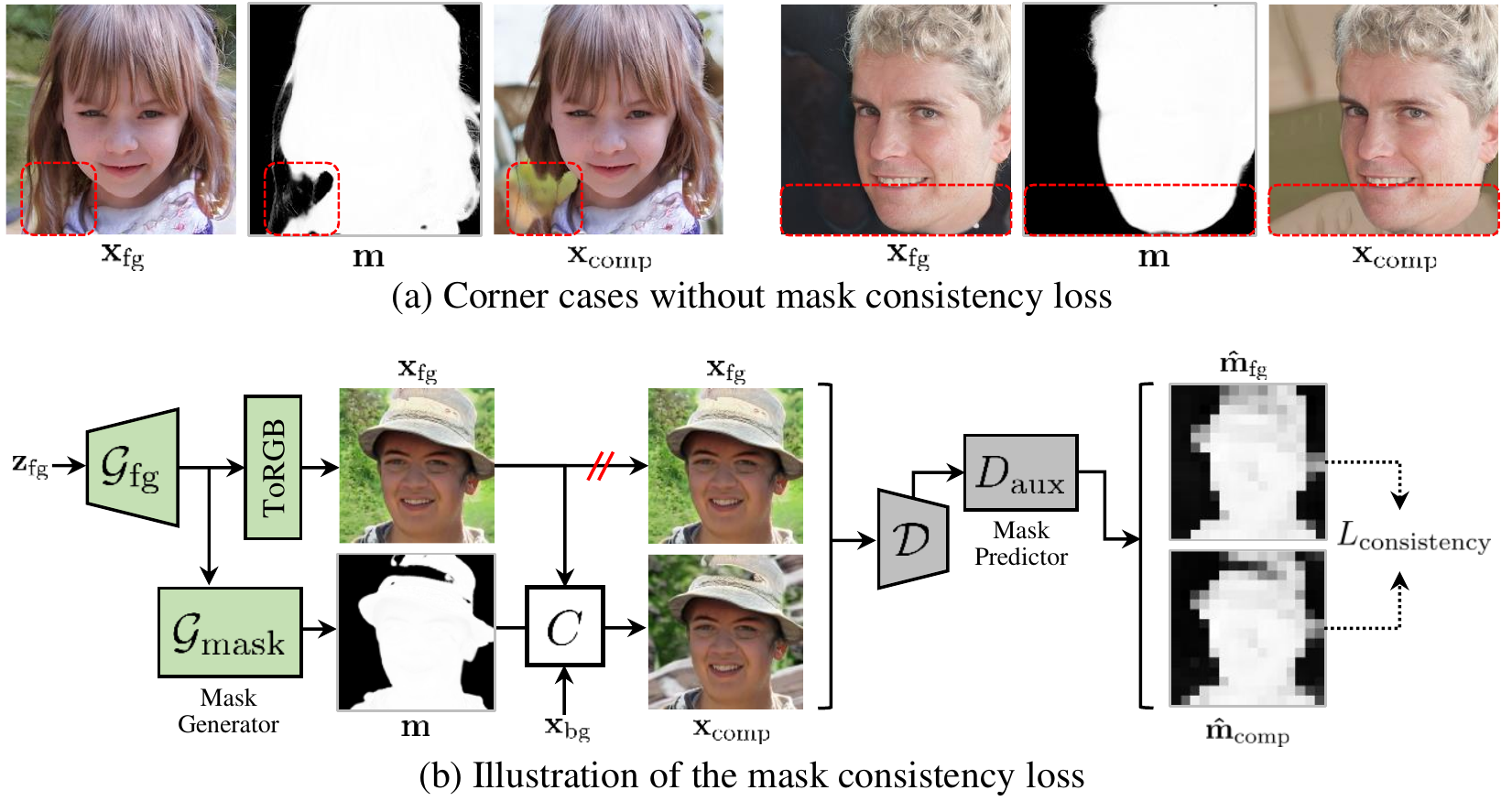}
    \caption{
    \textbf{Mask consistency loss.} 
    (a) Results without mask consistency loss show inconsistency between foreground images and composite images, e.g., cutting off long hair or adding shoulders. 
    (b) Mask consistency loss computes discrepancy between the predicted masks of the foreground and composite images.
    \textcolor{red}{//} denotes a stop gradient operator.
}
    \label{fig:mask_consistency}
\end{center}
\end{figure*}

\subsubsection{Mask consistency loss.} 
We observe that object regions of foreground $\xfg$ and composite $\xcomp$ can be inconsistent. For example, the mask may cut off the long hair in the foreground so that the composite image becomes a face with short hair. 
As another example, the missing body part of the foreground object may be supplemented from the background (both cases are shown in \fref{fig:mask_consistency}(a)).

Hence, we demand the mask predicted from the composite image to be consistent to the mask predicted from the foreground image (\fref{fig:mask_consistency}(b)):
\begin{equation} \label{eq:mask_consistency}
L_{\text{consistency}} = \frac{1}{|{\hat{\m}_{\text{comp}}|}} \| \daux(\sg(\xfg)) - \hat{\m}_{\text{comp}} \|^2_2,
\end{equation}
where $\hat{\m}_{\text{comp}}=\daux(\xcomp)$ and $\sg(\cdot)$ denotes a stop gradient operator.

As the mask predictor regresses the mask from the mask generator given a composite image, imposing consistency between the two masks encourages the foreground object region and the generated mask to resemble each other.

\subsubsection{Coarse mask loss.}
We adopt binarization loss and area loss following previous methods \cite{bielski2019emergence,abdal2021labels4free}. The binarization loss pushes the alpha values in the masks to either 0 or 1:
\begin{equation} \label{eq:binarize}
    L_\text{binary} = \mathbb{E} [\min(\mc, 1-\mc) ].
\end{equation}
The area loss penalizes the ratio of a mask being less than $\phi_1$ to promote using the foreground images more than $\phi_1$, i.e., preventing the background image from taking charge of everywhere, which is a degenerate solution:
\begin{equation} \label{eq:coarse}
L_\text{area}^\text{coarse} = \max(0, \phi_1 - \frac{1}{|\mc|} \sum \mc ),
\end{equation}
where $|\m|$ denotes the number of pixels in the mask image and $\phi_1$ is set to 0.35 for all experiments (unless otherwise noted).
The final coarse mask loss is:
\begin{equation}
L_{\mc} = L_\text{binary} + L_\text{area}^\text{coarse}.
\end{equation}

\subsubsection{Fine mask loss.}
The fine mask aims to capture details like hair, fur, and whiskers. Such a thin body becomes transparent due to the property of light. Hence, we do not use the binarization loss to free the masks to bear medium values between 0 and 1. Instead, we impose an inverse area loss to prevent the fine mask from taking charge of too large area:
\begin{equation} \label{eq:fineloss}
L_{\mf} = L_{\text{area}}^\text{fine} = \max(0, \phi_2 - \frac{1}{|\hmf|} \sum (1-\hmf) ),
\end{equation}
where $\hmf=\m-\mc$ to penalize the area where the fine mask actually contributes after clipping.
$\phi_2$ is set to 0.01 in all experiments. 
More details about the mask are illustrated in \aref{appendix:masks}.

\subsubsection{Background participation loss.}
We sometimes observe that the alpha mask tries to employ foreground images excessively. As a remedy, we penalize the difference between the composite image and the background image. It indirectly removes the excessive spread of the alpha mask.
\begin{equation} \label{eq:minimizefg}
L_{\text{reg}} =  \frac{1}{|\xcomp|} \| \xcomp - \xbg \|^2_2
\end{equation}
Intuitively, an easy way to reduce the difference between the composite image and the background is to remove unnecessary foreground areas that do not harm the realism of the composite image.

\subsubsection{Overall objective.}
Consequently, our full loss functions are:
\begin{equation}
L_\text{total}^\D = L_\text{adv}^\D + L_\text{R1}^\D + L_{\text{pred}},
\end{equation}
\begin{equation}
L_\text{total}^\G = L_{\text{adv}}^\G + L_\text{consistency} + \lambda_\text{coarse} L_{\mc} + \lambda_\text{fine} L_{\mf} + L_{\text{reg}}.
\end{equation}

\section{Experiments}

\subsection{Implementation Detail} \label{Implementation Detail}

Our foreground generator and background generator are based on StyleGAN2\cite{karras2020analyzing}. For simplicity, we remove output skip connections in the synthesis network and use a shallow mapping network \cite{karras2020training}. 
The foreground and background generators, $\gfg$ and $\gbg$, use a slightly modified StyleGAN2 structure. Number of channels of the latent codes and the feature maps in $\gfg$ and $\gbg$ become \textthreequarters$ $ and \textonequarter$ $, respectively.
As a result, the total number of parameters reduces by about half. 
Similar to \cite{yang2022learning}, background codes are shared with foreground codes. More precisely, we borrow the front part of the $\zfg\sim \mathcal{N}(0,I)$ and use it as $\zbg$.


We train our model on a single RTX-3090 for a period of about 100 hours. In all experiments, we trained our model for 300K iterations with a batch size of 16.
We follow training parameters from StyleGAN2 but do not use mixing regularization. 
Mask consistency loss and background participation loss update the model every other iteration. We set $\lambda_\text{coarse} = \lambda_\text{fine} = 5$. The coefficient of binarization loss is linearly reduced to 0.5 over the first 5K iterations.

\subsection{Setup}

\subsubsection{Datasets.}
We evaluate our model on FFHQ\cite{karras2019style} and AFHQv2-Cat\cite{choi2020starganv2,karras2021alias}.
FFHQ has 70,000 high-quality images of human faces. It has faces of various races and poses and also has good coverage of accessories such as eyeglasses, hats, etc. AFHQv2-Cat contains 5000 images of cat faces.
The rebuilt (v2) dataset has higher quality due to proper resizing and compression. We also trained our model on unaligned datasets such as LSUN-Object \cite{yu2015lsun}, and CUB\cite{wah2011caltech} (see \aref{appendix:unaligned} for details and results). 
All models are trained at 256$\times$256 resolution.

\subsubsection{Pseudo ground truth masks.}
We evaluate the generated mask quality to show foreground-background separation performance. 
Because the generated images do not contain ground truth masks for evaluating the generated foreground masks, we adopt TRACER \cite{lee2021tracer} to prepare pseudo ground truth masks. It provides detailed masks including hair and whiskers, which are not captured by segmentation networks used in PSeg \cite{bielski2019emergence} and Labels4Free \cite{abdal2021labels4free}.
Please refer to \aref{appendix:gt} for their comparison.

\subsubsection{Metrics.}
To quantitatively measure the quality of images, we compute Fr\'echet Inception Distance (FID)\cite{heusel2017gans} between generated foreground images and all training images. Unless otherwise specified, all results were obtained with 50,000 generated images following \cite{karras2020training,karras2021alias}.
To quantitatively measure the quality of masks, we employ intersection over union (IoU) for the foreground and background, and their mean (mIoU). IoU and mIoU measure the overlap between prediction masks and ground truth masks. Furthermore, we report standard segmentation metrics: precision, recall, F1 score, and segmentation accuracy following \cite{abdal2021labels4free}.

\begin{figure*}[t]
\begin{center}
    \includegraphics[width=0.85\linewidth]{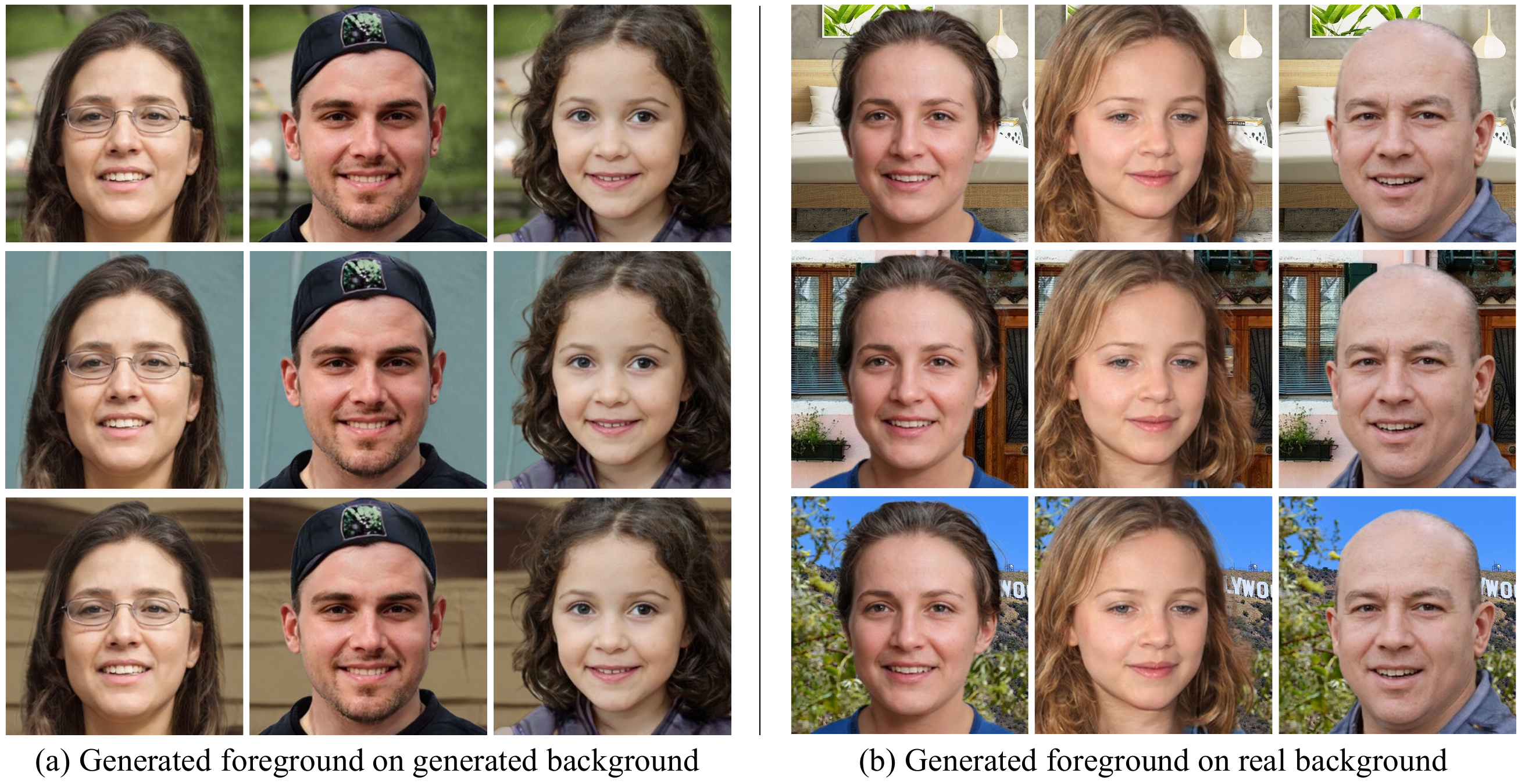}
    \caption{
    \textbf{Composite images.} The same person is placed on the vertical axis, and the same background is placed on the horizontal axis. 
    }
    \label{fig:composite}
\end{center}
\end{figure*}

\begin{figure*}[t]
\begin{center}
    \includegraphics[width=\linewidth]{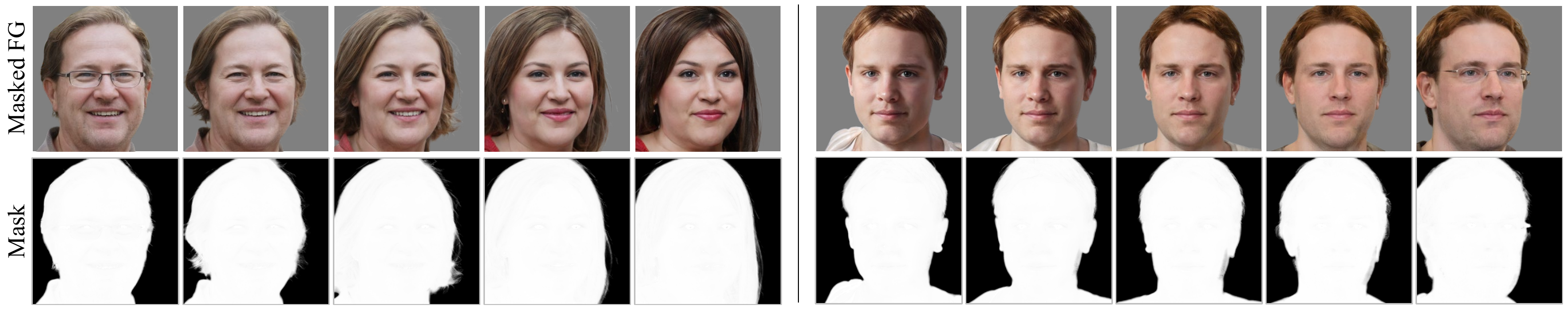}
    \caption{
    \textbf{Latent space interpolation.} We show the mask changes naturally as the image changes.
    }
    \label{fig:interpolation}
\end{center}
\end{figure*}

\subsection{Experiments about masked foregrounds}
Thanks to the high quality mask generated by our model, the masked foreground object can be naturally combined with various backgrounds, as shown in \fref{fig:composite}. We sample random background latent codes for the backgrounds used in \fref{fig:composite} (a). 
\fref{fig:interpolation} shows that the interpolation in the foreground latent space not only changes the image but also changes the shape of masks correspondingly.

\subsection{Ablation study}

\begin{figure}[t]
\begin{center}
    \includegraphics[width=0.9\linewidth]{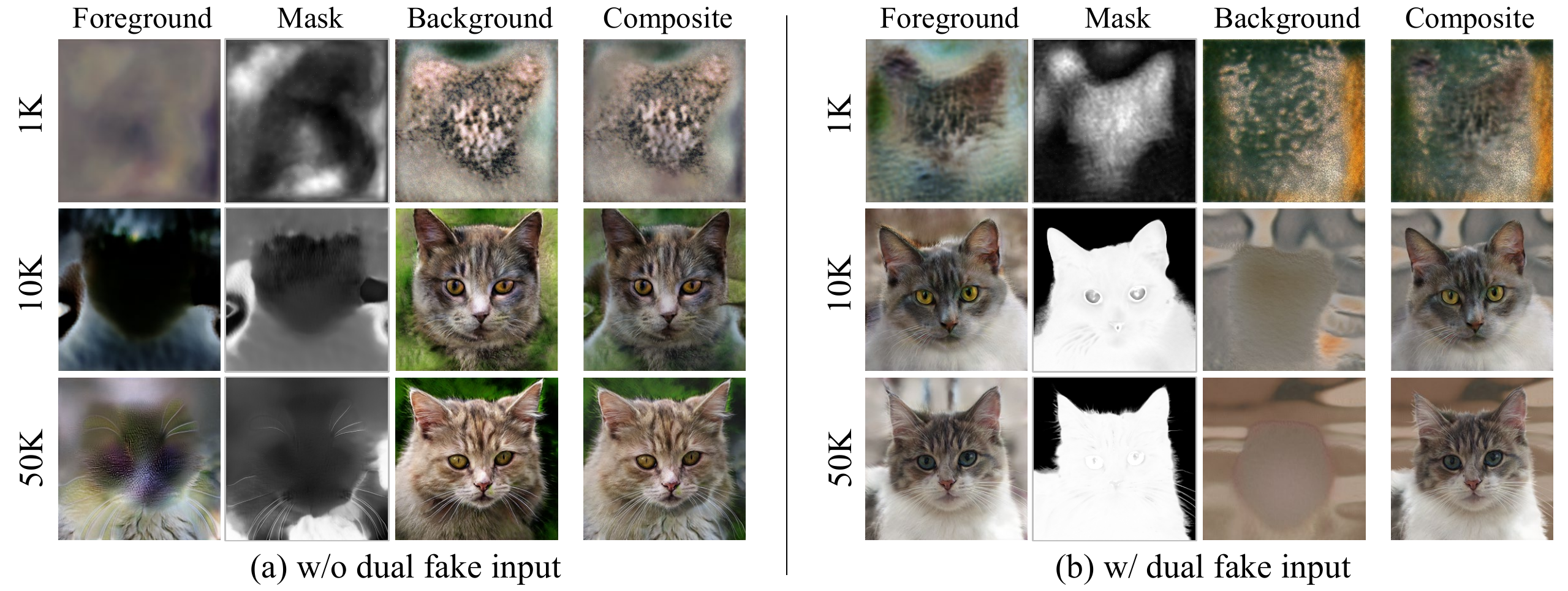}
    \caption{\textbf{Ablation of dual fake input strategy.} Each row shows the early results as training proceeds. (a)without dual fake input, foreground object has shown in background image. (b)With dual fake input, it is separated naturally.}
    \label{fig:nodual}
    
\end{center}
\end{figure}

\begin{figure*}[t]
\begin{center}
    \includegraphics[width=0.9\linewidth]{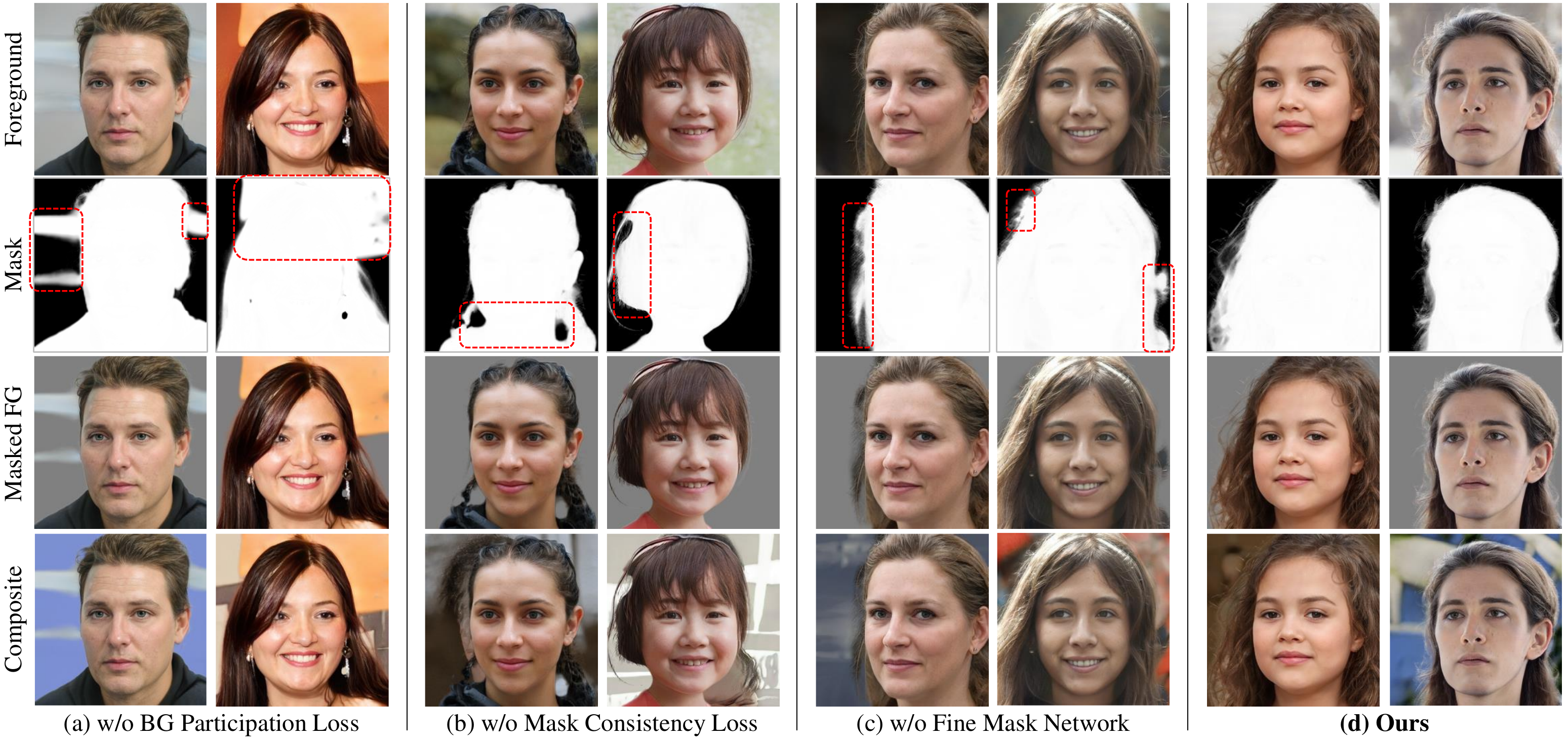}
    \caption{\textbf{Ablation of our methods.} Three columns show the result without one component. (a) The masks have background area. (b) The masks are not align correctly with the foreground object region. (c) The masks bring the surrounding background. (d) our method produces fine-grained masks.}
    \label{fig:ablation}
\end{center}
\end{figure*}

\subsubsection{Dual fake input strategy.}

\fref{fig:nodual}(a) shows that the foreground and mask generator fail to synthesize meaningful foreground and mask, respectively, without the dual fake input strategy. We suppose that it is easier for the generators to focus on the background to synthesize realistic composite images because the foreground generator has a more complicated task: producing foreground images and the masks with more parameters. On the other hand, with the dual fake input strategy, the foreground images have clear objects as they should resemble the training images (\fref{fig:nodual}(b)).

\subsubsection{Ablation of losses.}
\fref{fig:ablation} visually compares the results without one component at a time and our full method. Without background participation loss (\eref{eq:minimizefg}), the masks tend to be wider than the foreground object area.
Without mask consistency loss (\eref{eq:mask_consistency}), the mask does not align correctly with the foreground object region.
Without the fine mask network, the fine details in the mask tend to be less accurate, 
especially on the region between hair and background. With all components combined, our method produces fine masks aligned to the foreground object region.

\tref{tab:ablation} provides quantitative ablation study. Decrease in quality of masks show the necessity of background participation loss and mask consistency loss. Ablating any of the components harms FID, implying that spatial understanding in the generators is important for the quality of images. We suppose that the influence of the fine mask generator is negligible in the metrics for the masks because the metrics are not sensitive enough to reflect changes in small area.

\begin{table}[t]
\caption{ Quantitative comparison of ablation study on FFHQ. }
\centering
\renewcommand{\arraystretch}{1.2}
\resizebox{0.8\columnwidth}{!}{%
\begin{tabular}{l|ccccccc}
\multicolumn{1}{l|}{Setting}  & IoU(fg/bg)   & mIoU & recall  & precision  & F1   & Accuracy   & FID   \\ \hline
w/o Fine Mask Network & \textbf{0.93/0.83}   & \textbf{0.88} & 0.95 & \textbf{0.98} & \textbf{0.96} & \textbf{0.93} & 9.48 \\
w/o BG Participation   & 0.91/0.77 & 0.84 & \textbf{0.97} & 0.94 & 0.95 & 0.91 & 9.79 \\
w/o Mask Consistency & 0.92/0.81 & 0.86 & 0.93 & 0.98 & \textbf{0.96} & 0.92 & 9.53 \\ \hline
Full ours   & \textbf{0.93}/0.82  & \textbf{0.88} & 0.95 & \textbf{0.98} & \textbf{0.96} & \textbf{0.93} & \textbf{8.72}
\end{tabular}
}
\label{tab:ablation}
\end{table}

\subsection{Comparisons}
\subsubsection{Competitors.}
We choose PSeg\footnote{https://github.com/adambielski/perturbed-seg} \cite{bielski2019emergence} and Labels4Free\footnote{https://github.com/RameenAbdal/Labels4Free} (L4F in short, \cite{abdal2021labels4free}) as our competitors. For a fair comparison, we trained L4F in FFHQ and AFHQ under the same conditions, i.e., batch size, data augmentations, training iterations. We pretrain StyleGAN2\footnote{https://github.com/rosinality/stylegan2-pytorch} for $256\times256$ resolution and then train the alpha network with its official setting. As Labels4Free does not conduct experiments on AFHQ, we train their alpha network for the same number of iterations on FFHQ (=1K), and manually find the working hyperparameters: $\lambda_2=3$ and $\phi_2=0.2$\footnote{Without setting $\phi_2$, all masks of Labels4Free saturate to 1.}. For PSeg, we add additional layers to their networks for $256\times256$ resolution since PSeg conducts their experiments in $128\times128$ resolution. When training Pseg, we followed the default setting reported in the paper. We do not include FineGAN \cite{singh2019finegan} because it does not focus on foreground-background separation and it requires an external pretrained object detector for supervision.


\begin{figure}[t]
\begin{center}
    \includegraphics[width=0.9\linewidth]{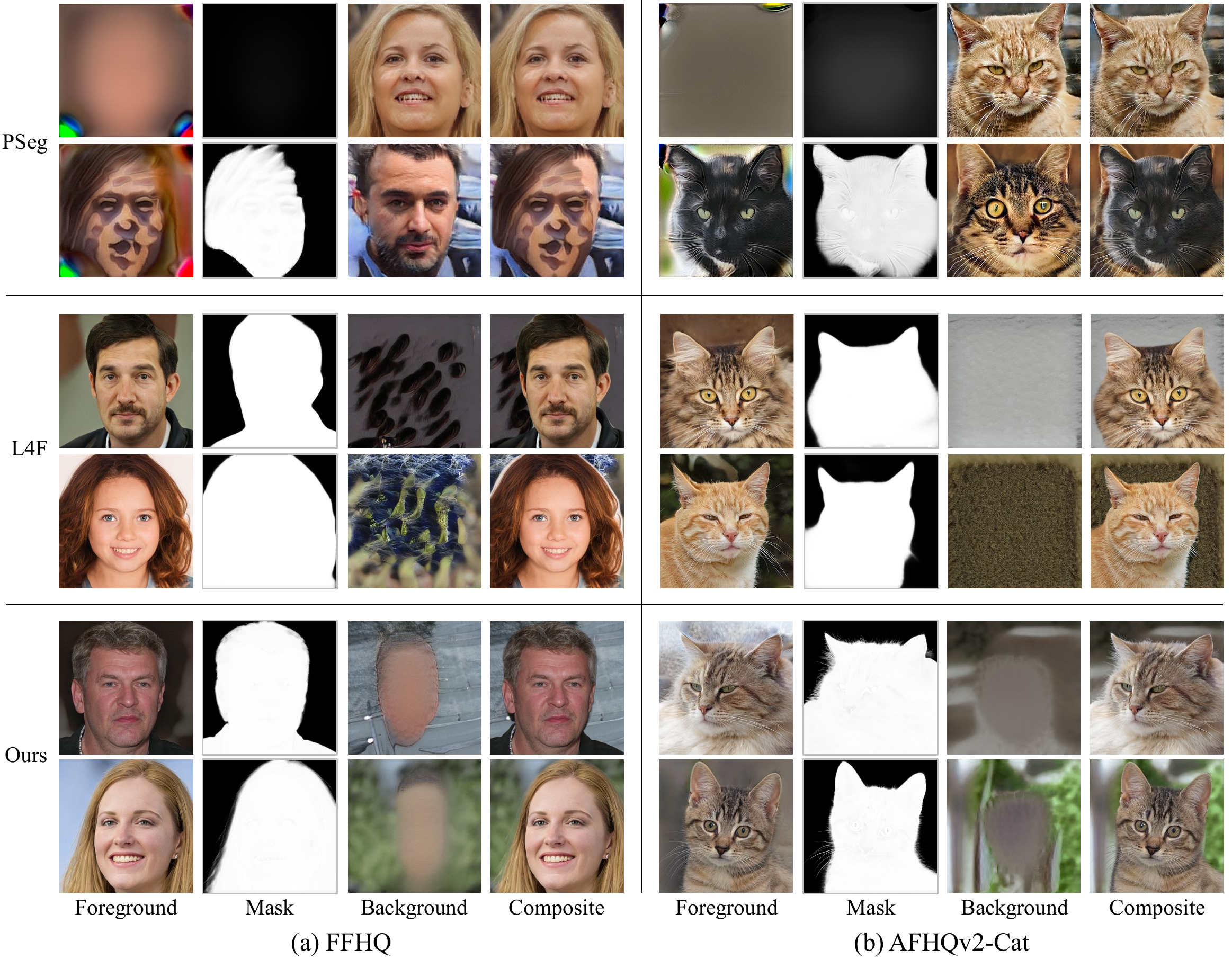}
    \caption{
    Qualitative comparison of image composition results on FFHQ and AFHQv2-Cat. 
    }
    \label{fig:comparison}
\end{center}
\end{figure}

\subsubsection{Qualitative Results.}
\fref{fig:comparison} provides a qualitative comparison between the methods.
PSeg rarely succeeds in synthesizing proper foreground images and mostly draws objects on the background images. We suppose the reason to be the unmet assumption: for faces, the foreground cannot help touching the edges, thus shifting the foreground will not be realistic. Labels4Free somewhat successes separating foregrounds. However, their masks are not accurate enough and the composition leaves artifacts on the boundaries. 
In contrast, our method produces masks that accurately capture the foreground object, even including hair, fur, and whiskers. Uncurated samples can be found in \aref{appendix:uncurated}

\subsubsection{Quantitative Results.}
\tref{tab:compare} reports how well the masks align with the foreground object region. The pseudo ground truth masks are obtained by feeding the foreground images to TRACER \cite{lee2021tracer}. Our method consistently outperforms the competitors in all settings: different levels of truncation and datasets.
\aref{appendix:quant} provides the results with other choices of pseudo ground truth.

\tref{tab:fid} quantitatively compares the visual quality of the generated images. Our method achieves FIDs comparable to Labels4Free whose foreground generator equals the pretrained StyleGAN2 while drastically improving the masks. 

\begin{table}[t]
\caption{Quantitative comparison of alpha mask results on FFHQ and AFHQv2-Cat. We report the result with/without truncation($\psi$=1.0, 0.7) and the threshold for the mask is 0.5(Ours, PSeg) and 0.9(L4F). }
\centering
\setlength\tabcolsep{3.8pt}
\renewcommand{\arraystretch}{1.2}
\resizebox{0.8\columnwidth}{!}{%
\begin{tabular}{c|c|c|cccccc}
 & $\psi$ & method & IoU(fg/bg) & mIoU & recall & precision & F1 & Accuracy \\ \hline
\multirow{6}{*}{FFHQ} & \multirow{3}{*}{1.0} & PSeg & 0.05/0.23 & 0.14 & 0.05 & 0.18 & 0.07 & 0.05 \\
 &  & L4F & 0.87/0.70 & 0.78 & 0.92 & 0.94 & 0.93 & 0.87 \\
 &  & Ours & \textbf{0.93/0.82} & \textbf{0.88} & \textbf{0.95} & \textbf{0.98} & \textbf{0.96} & \textbf{0.93} \\ \cline{2-9} 
 & \multirow{3}{*}{0.7} & PSeg & 0.01/0.23 & 0.12 & 0.01 & 0.04 & 0.01 & 0.01 \\
 &  & L4F & 0.91/0.79 & 0.85 & 0.94 & 0.97 & 0.95 & 0.91 \\
 &  & Ours & \textbf{0.95/0.88} & \textbf{0.91} & \textbf{0.95} & \textbf{0.99} & \textbf{0.97} & \textbf{0.95} \\ \hline
\multirow{6}{*}{AFHQv2-Cat} & \multirow{3}{*}{1.0} & PSeg & 0.06/0.23 & 0.15 & 0.06 & 0.16 & 0.07 & 0.06 \\
 &  & L4F & 0.91/0.80 & 0.86 & 0.93 & \textbf{0.98} & 0.95 & 0.91 \\
 &  & Ours & \textbf{0.94/0.82} & \textbf{0.88} & \textbf{0.98} & 0.96 & \textbf{0.97} & \textbf{0.94} \\ \cline{2-9} 
 & \multirow{3}{*}{0.7} & PSeg & 0.01/0.19 & 0.10 & 0.01 & 0.12 & 0.01 & 0.01 \\
 &  & L4F & 0.91/0.79 & 0.85 & 0.94 & \textbf{0.97} & 0.95 & 0.91 \\
 &  & Ours & \textbf{0.95/0.87} & \textbf{0.91} & \textbf{0.98} & \textbf{0.97} & \textbf{0.97} & \textbf{0.95}
\end{tabular}
}

\label{tab:compare}
\end{table}
\begin{table}[t]
\caption{ Quantitative comparison of generated foreground images on FFHQ and AFHQv2-Cat. Foreground generator of L4F equals to the pretrained StyleGAN2.}
\centering
\setlength\tabcolsep{6pt}
\renewcommand{\arraystretch}{1.2}
\resizebox{0.5\columnwidth}{!}{%
\begin{tabular}{l|cc}
 & \multicolumn{2}{c}{FID} \\ \cline{2-3} 
 & \multicolumn{1}{c|}{FFHQ} & AFHQv2-Cat \\ \hline
Pseg & \multicolumn{1}{c|}{62.44} & 12.71 \\ \hline
\begin{tabular}[c]{@{}l@{}}Labels4Free \scriptsize{(=StyleGAN2)}\end{tabular} & \multicolumn{1}{c|}{\textbf{6.51}} & \textbf{5.19} \\ \hline
Ours & \multicolumn{1}{c|}{8.72} & 6.34
\end{tabular}
}
\label{tab:fid}
\end{table}

\subsection{Segmenting real images.}
In addition, we demonstrate an extension of our method for segmenting real images. Following Labels4Free, we use 1K images and their ground truth segmentation masks from CelebAMask-HQ dataset \cite{CelebAMask-HQ} for evaluation. We employ the original inversion method from StyleGAN2. While \tref{tab:inversion} shows that our method achieves similar performance, \fref{fig:inversion} shows that our method produces much more accurate and finer masks.

\begin{table}[t]
\caption{Quantitative comparison of alpha masks from inverted real images on CelebAMask-HQ. We report the result with the original inversion method from StyleGAN2 and the threshold for the mask is 0.5(Ours) and 0.9(Labels4Free).}
\renewcommand{\arraystretch}{1.5}
\centering
\resizebox{0.6\columnwidth}{!}{%
\begin{tabular}{l|cccccc}
 & IoU(fg/bg) & \multicolumn{1}{c}{mIoU} & \multicolumn{1}{c}{recall} & \multicolumn{1}{c}{precision} & \multicolumn{1}{c}{f1} & \multicolumn{1}{c}{accuracy} \\ \hline
Labels4Free & \textbf{0.93/0.81} & \textbf{0.87} & \textbf{0.97} & 0.95 & \textbf{0.96} & \textbf{0.93} \\
Ours & 0.92/\textbf{0.81} & \textbf{0.87} & 0.95 & \textbf{0.97} & \textbf{0.96} & 0.92
\end{tabular}
}
\label{tab:inversion}
\end{table}

\begin{figure}[!t]
\begin{center}
    \includegraphics[width=0.9\linewidth]{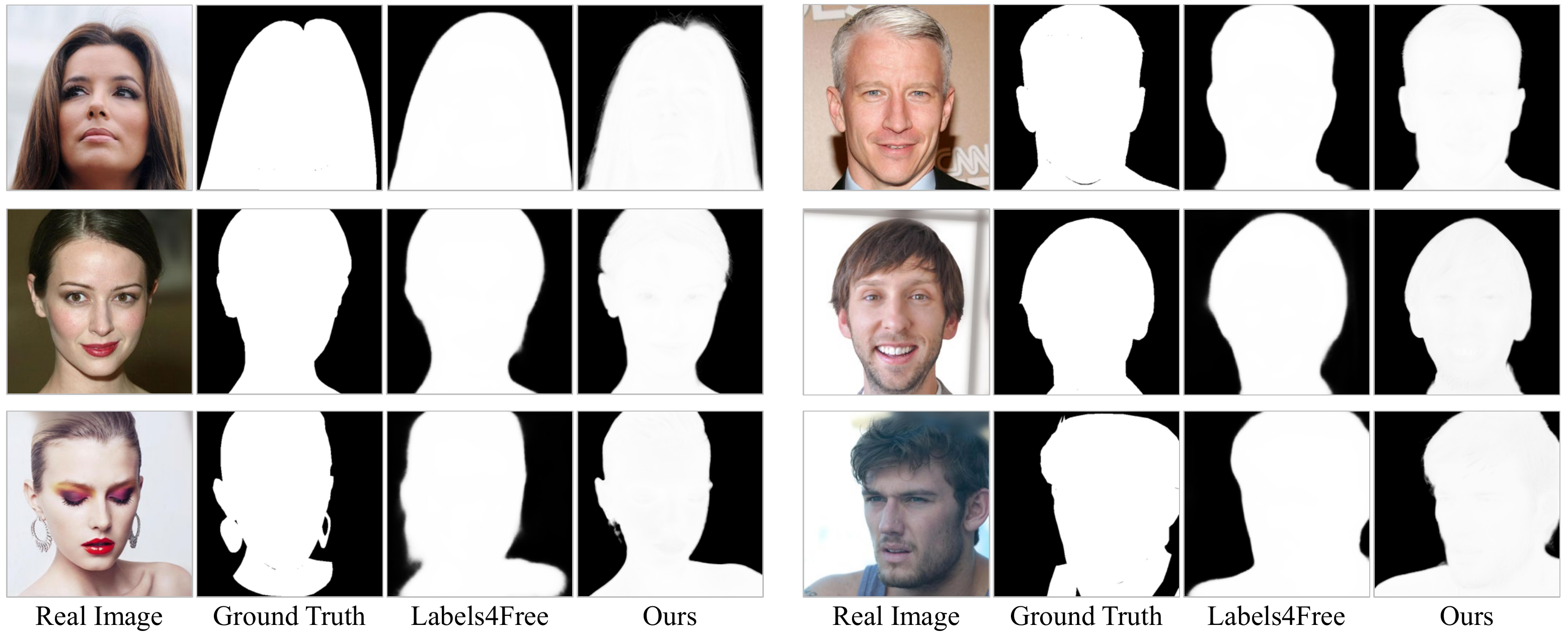}
    \caption{
    Visual comparison on segmenting real images.}
    \label{fig:inversion}
    
\end{center}
\end{figure}

\begin{figure*}[!t]
\begin{center}
    \includegraphics[width=0.9\linewidth]{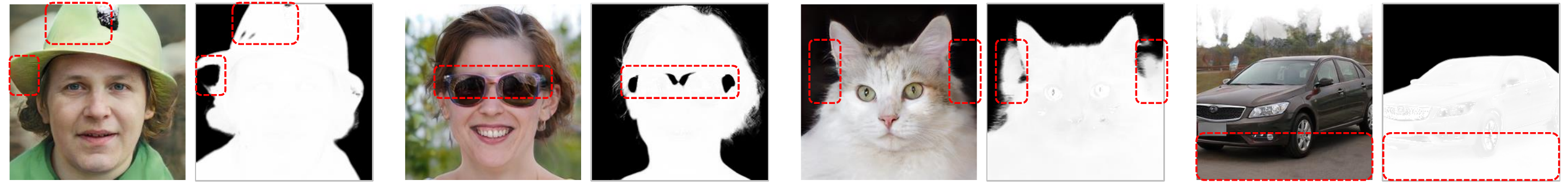}
    \caption{
    Various kinds of failures in our model (foreground-mask pair).
    }
    \label{fig:fail}
\end{center}
\end{figure*}


\section{Conclusion and discussion}
Understanding spatial semantics in the synthesized images is an important research problem in GANs. In this paper, we proposed a GAN framework for foreground-aware image synthesis, generating images as a combination of foreground and background according to a mask. Our method achieves dramatic improvement in the fine details of the masks without any supervision or dataset-tailored assumption. Our model also can be trained on unaligned datasets such as LSUN, indicating that our method generally works well. 

However, we observe exceptional cases where the mask generator struggles in \fref{fig:fail}. We suggest one of the main reasons to be the ambiguity of the task itself. In CompCars \cite{yang2015large}, the road below the vehicles is often marked as foreground. It is a reasonable choice because the road is physically close to the vehicles. 
Using a minimal amount of human supervision for resolving such ambiguity would be a sensible research direction, e.g., specifying foreground or background by scribbles on one or a few images.
In some cases, the mask misses a small portion of the object area. This might be because the composite image is natural enough, even if the mask is inappropriate.
We hope that our success in the common datasets in GAN literature sheds light on foreground-aware image synthesis.

\subsubsection{Acknowledgement}
This work was supported by the National Research Foundation of Korea(NRF) grant (No. 2022R1F1A107624111) funded by the Korea government (MSIT).

\clearpage
%
%
\bibliographystyle{splncs04}
\bibliography{egbib}

\clearpage
\else

\fi
\begin{appendix}
\title{Supplementary Materials for \\ FurryGAN: High Quality Foreground-aware Image Synthesis}
\titlerunning{ECCV-22 submission ID \ECCVSubNumber} 
\authorrunning{ECCV-22 submission ID \ECCVSubNumber} 
\author{Jeongmin Bae \and
Mingi Kwon \and
Youngjung Uh\thanks{Corresponding author}}
\authorrunning{J. Bae et al.}
%
\institute{Yonsei University\\
\email{\{jaymin.bae, kwonmingi, yj.uh\}@yonsei.ac.kr}}
\maketitle

\renewcommand{\thetable}{S\arabic{table}}
\renewcommand{\thefigure}{S\arabic{figure}}

We provide the following supplementary materials:
\begin{enumerate}[label=\Alph*]
    \item Uncurated visual comparison of Labels4Free and ours
    \item Choice of pseudo ground truth masks
    \item Quantitative results with alternative pseudo ground truth masks
    \item Examples of style mixing
    \item Additional details and visualization of the mask
    \item Results on unaligned datasets (LSUN-Church, LSUN-Horse, and CUB)
    \item User study on mask quality (between Lables4Free and ours)
\end{enumerate}



\section{Uncurated comparison}
\label{appendix:uncurated}
\fref{fig:uncurated_ffhq_07}-\ref{fig:uncurated_afhq_07} (located at the end for clear spacing) present uncurated visual comparisons between Labels4Free and ours on FFHQ and AFHQv2-Cat. The columns represent foreground images, alpha masks, and composite images with generated backgrounds.
While Labels4Free often misses clothes and whiskers, our method produces more accurate and detailed masks, especially on hair, fur, and whiskers. Consistency between the generated masks and the actual foreground region in the composite image also demonstrates the superiority of our method.

\section{Choice of pseudo ground truth masks}
\label{appendix:gt}
\begin{figure}[t]
\begin{center}
    \includegraphics[width=1\linewidth]{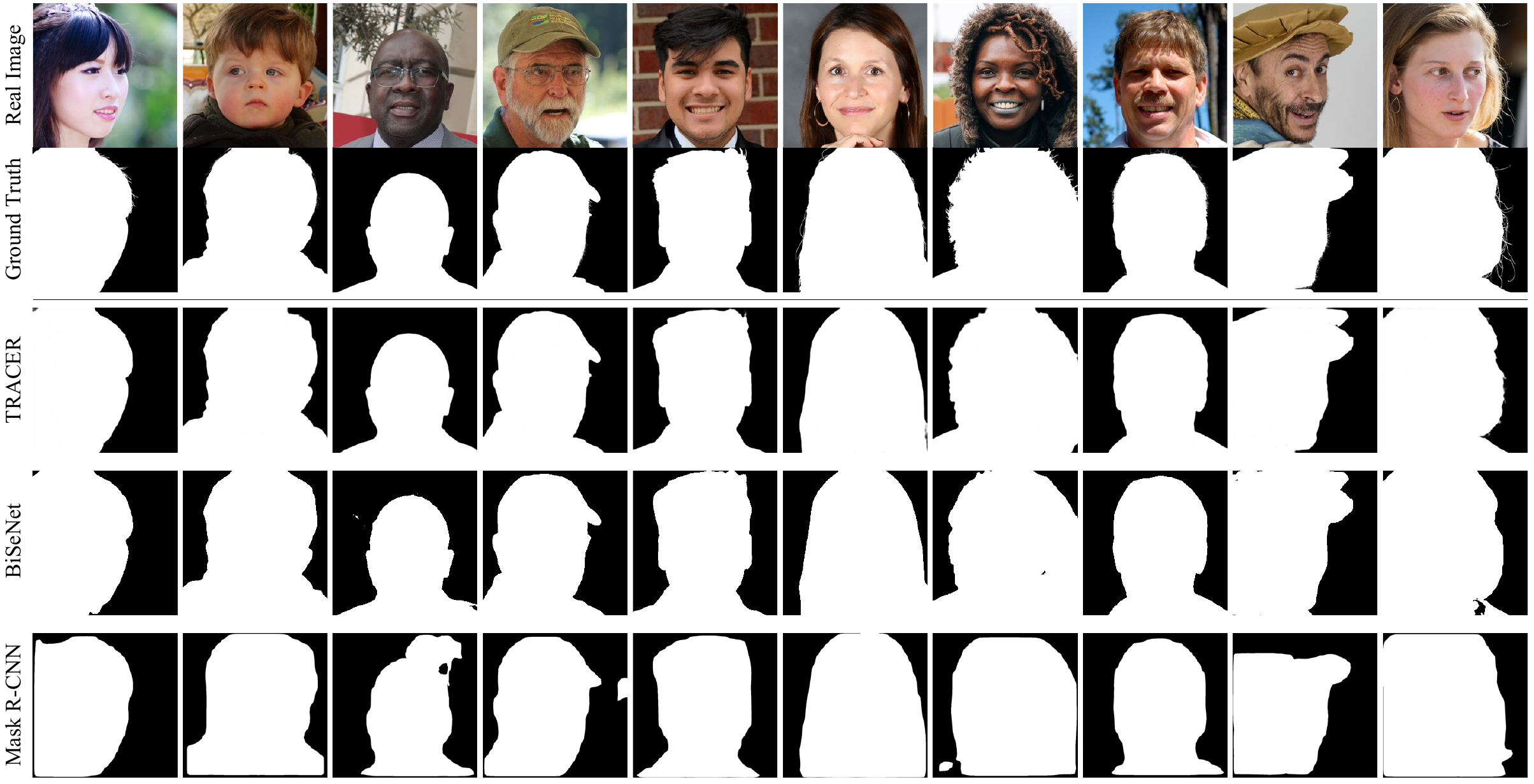}
    \caption{
    \textbf{Qualitative comparison of masks.} We manually annotated ground truth masks (second row). TRACER produces masks very similar to the ground truth. BiSeNet also shows acceptable performance, but it often misclassifies the background as a foreground (3rd, 9th column) and vice versa (10th column). Mask R-CNN is relatively poor in quality, especially near the borders of the mask.
    }
    \label{fig:gt}
    \vspace{-10pt}
\end{center}
\end{figure}

In this section, we provide the grounds for choosing TRACER (TE7) \cite{lee2021tracer} to prepare pseudo ground truth masks over BiSeNet \cite{yu2018bisenet} (in Labels4Free \cite{abdal2021labels4free}) and Mask R-CNN\footnote{https://github.com/facebookresearch/maskrcnn-benchmark} (in PSeg \cite{bielski2019emergence}).
As FFHQ do not have ground truth masks, we manually annotate ten images for the evaluation. The images are broadly chosen to cover various ages, genders, ethnic groups, and accessories. \fref{fig:gt} shows the chosen images, annotated ground truths, and the pseudo ground truths from the methods. The quantitative comparison also reveals that TRACER achieves the best performance. Note that CelebAMask-HQ does not suffice to serve as the benchmark because BiSeNet is trained on it.


\begin{table}[!ht]
\setlength\tabcolsep{4pt}
\renewcommand{\arraystretch}{1.2}
\centering
\begin{tabular}{c|cccccc}
method & IoU(fg/bg) & \multicolumn{1}{c}{mIoU} & \multicolumn{1}{c}{recall} & \multicolumn{1}{c}{precision} & \multicolumn{1}{c}{F1} & \multicolumn{1}{c}{Accuracy} \\ \hline
Mask R-CNN & 0.92/0.85 & 0.88 & 0.97 & 0.94 & 0.96 & 0.92 \\
BiSeNet & 0.98/0.96 & 0.97 & 0.99 & \textbf{0.99} & \textbf{0.99} & 0.98 \\
TRACER & \textbf{0.99/0.97} & \textbf{0.98} & \textbf{1.00} & \textbf{0.99} & \textbf{0.99} & \textbf{0.99}
\end{tabular}
\caption{\textbf{Quantitative comparison of predicted masks on the ten selected FFHQ images.} We evaluate the performance of the models with ten manually annotated ground truth masks.}
\label{tab:tracer}
\end{table}

\begin{figure}[!ht]
\begin{center}
    \includegraphics[width=1\linewidth]{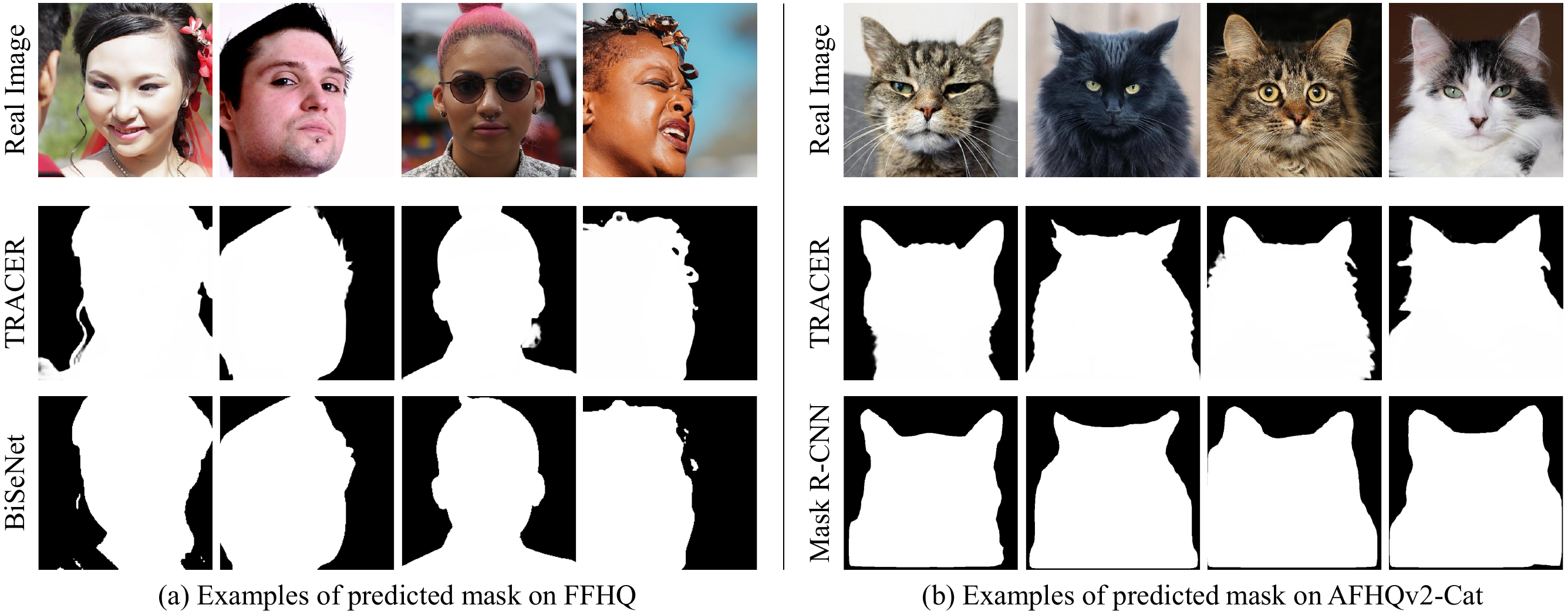}
    \caption{
    \textbf{Further comparison of TRACER and other methods.} We evaluate each model on real images from FFHQ and AFHQv2-Cat datasets. 
    }
    \label{fig:gt_comparison}
\end{center}
\end{figure}

\fref{fig:gt_comparison} further contrast the performance of the methods. On FFHQ, TRACER captures even hair while BiSeNet struggles. On AFHQv2-Cat, TRACER precisely captures even long fur on the ears and the top of the heads. 

\begin{table}[!h]
\centering
\setlength\tabcolsep{3.8pt}
\renewcommand{\arraystretch}{1.2}
\begin{tabular}{c|c|c|cccccc}
 & $\psi$ & method & IoU(fg/bg) & mIoU & recall & precision & F1 & Accuracy \\ \hline
\multirow{6}{*}{\begin{tabular}[c]{@{}c@{}}FFHQ\\ (BiSeNet)\end{tabular}} & \multirow{3}{*}{1.0} & PSeg & 0.05/0.24 & 0.14 & 0.05 & 0.16 & 0.07 & 0.05 \\
 &  & L4F & 0.86/0.70 & 0.78 & 0.93 & 0.92 & 0.92 & 0.86 \\
 &  & Ours & \textbf{0.92/0.80} & \textbf{0.86} & \textbf{0.95} & \textbf{0.96} & \textbf{0.95} & \textbf{0.92} \\ \cline{2-9} 
 & \multirow{3}{*}{0.7} & PSeg & 0.01/0.23 & 0.12 & 0.01 & 0.04 & 0.01 & 0.01 \\
 &  & L4F & 0.94/0.87 & 0.91 & \textbf{0.96} & \textbf{0.99} & \textbf{0.97} & 0.94 \\
 &  & Ours & \textbf{0.95/0.89} & \textbf{0.92} & \textbf{0.96} & \textbf{0.99} & \textbf{0.97} & \textbf{0.95} \\ \hline
\multirow{6}{*}{\begin{tabular}[c]{@{}c@{}}AFHQv2-Cat\\ (Mask R-CNN)\end{tabular}} & \multirow{3}{*}{1.0} & PSeg & 0.06/0.21 & 0.13 & 0.06 & 0.17 & 0.07 & 0.06 \\
 &  & L4F & 0.88/\textbf{0.72} & 0.80 & 0.91 & \textbf{0.97} & 0.94 & 0.88 \\
 &  & Ours & \textbf{0.91/0.72} & \textbf{0.81} & \textbf{0.95} & 0.95 & \textbf{0.95} & \textbf{0.91} \\ \cline{2-9} 
 & \multirow{3}{*}{0.7} & PSeg & 0.01/0.17 & 0.09 & 0.01 & 0.13 & 0.01 & 0.01 \\
 &  & L4F & 0.91/\textbf{0.77} & \textbf{0.84} & 0.92 & \textbf{0.98} & 0.95 & 0.91 \\
 &  & Ours & \textbf{0.92/0.77} & \textbf{0.84} & \textbf{0.95} & 0.96 & \textbf{0.96} & \textbf{0.92}
\end{tabular}
\caption{\textbf{Quantitative comparison of alpha masks on FFHQ and AFHQv2-Cat.} We use results of BiSeNet trained on CelebAMask-HQ as ground truth for FFHQ and results of Facebook's Detectron2 Mask R-CNN Model (R101-FPN) as ground truth for AFHQv2-Cat. We report the result with/without truncation trick ($\psi$=0.7, 1.0). The threshold for the alpha mask is 0.5 in ours and PSeg, and 0.9 in Labels4Free.}

\label{tab:compare2}
\vspace{-10pt}
\end{table}

\section{Quantitative evaluation with alternative pseudo ground truth masks}
\label{appendix:quant}

In this section, we report quantitative results with other choices of generating pseudo ground truth masks: BiSeNet for FFHQ and Mask R-CNN for AFHQv2-Cat following Labels4Free\footnote{Labels4Free uses Mask R-CNN for LSUN-Cat.}.
\tref{tab:compare2} confirms the same rankings as the ones with TRACER; our method consistently outperforms the competitors in all settings.

\section{Style mixing}
\label{appendix:mixing}
\begin{figure*}[!h]
\begin{center}
    \includegraphics[width=0.8\linewidth]{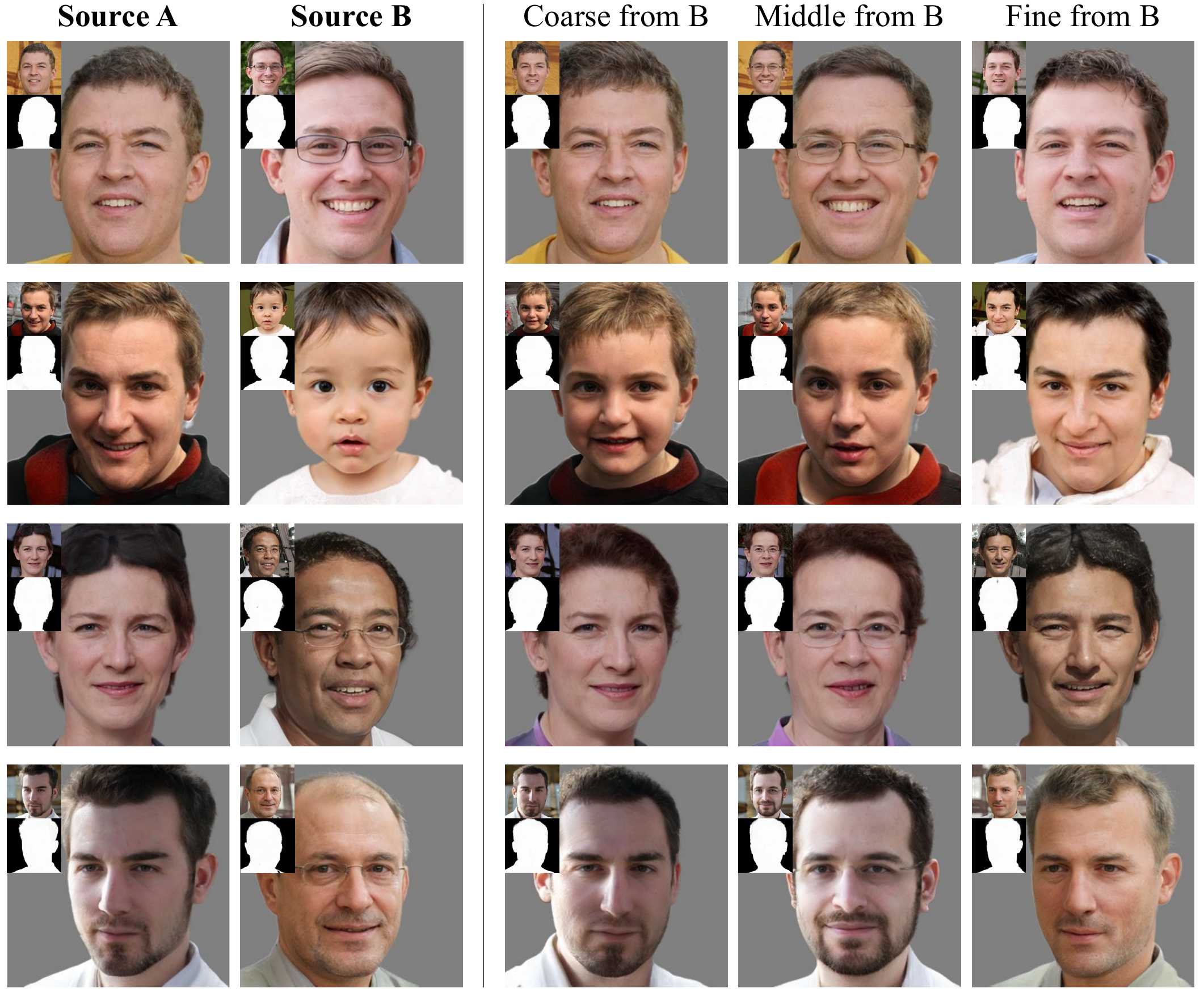}
    \caption{
    The two leftmost columns are source images denoted by A and B. The right side of the figure is the result of using the latent code of B instead of the latent code of A in the coarse ($4^2$-$8^2$), middle ($16^2$-$32^2$), and fine ($64^2$-$256^2$) layers, respectively. We demonstrate masked foreground images to show the changes in the foreground mask according to different style mixing. In addition, we provide the composite image and mask in the upper left corner of each image.}
    \label{fig:mixing}
\end{center}
\end{figure*}

Our generator supports style mixing since it is based on StyleGAN2. As coarse style affects shape in StyleGAN2, the masks of the coarse source determine the masks of the mixed results in our generator (\fref{fig:mixing}). Note that we do not use mixing regularization during the training.

\begin{figure*}[!h]
\vspace{-5pt}
\begin{center}
    \includegraphics[width=0.8\linewidth]{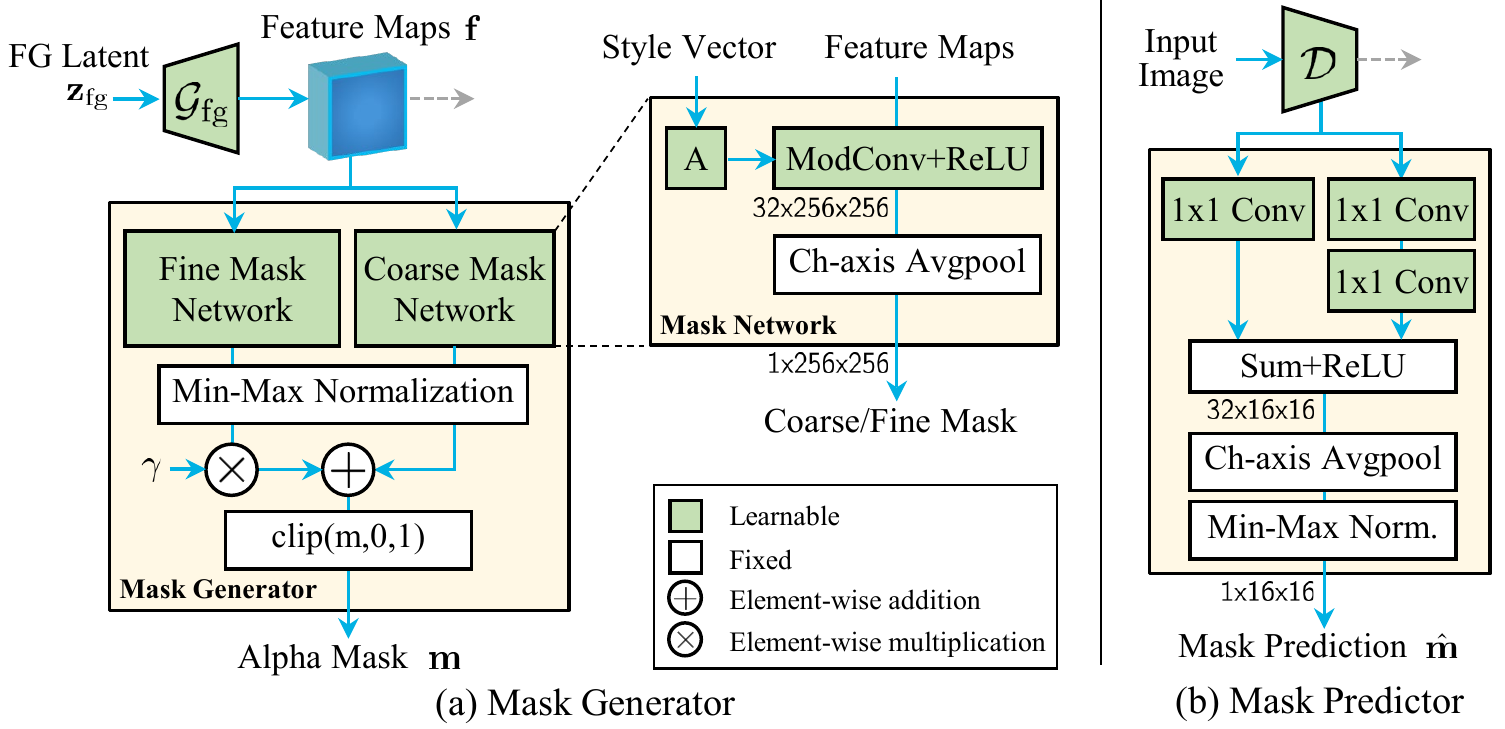}
    \caption{\textbf{Architecture of the mask generator and the mask predictor.} Coarse and fine mask networks use the same structure shown in the upper right corner of (a). $\gamma$ is defined in \eref{eq:final_mask}. For brevity, we omit the LeakyReLU activation function between the convolution layers of the right branch in (b).}
    \label{fig:gm_daux}
    \vspace{-10pt}
\end{center}
\end{figure*}

\section{
Details about masks
}
\label{appendix:masks}

In this section, we present the motivation for introducing fine masks and show additional mask visualizations. We assumed that the binarization loss (\eref{eq:binarize}) makes it difficult for the model to learn the matting-like details in the mask. These fine details are expected to occupy only a small part around the object boundary. Accordingly, we do not use binarization loss for the fine masks and use a very low threshold value for the inverse area loss (\eref{eq:fineloss}).

\begin{figure}[!ht]
\vspace{10pt}
\begin{center}
    \includegraphics[width=0.85\linewidth]{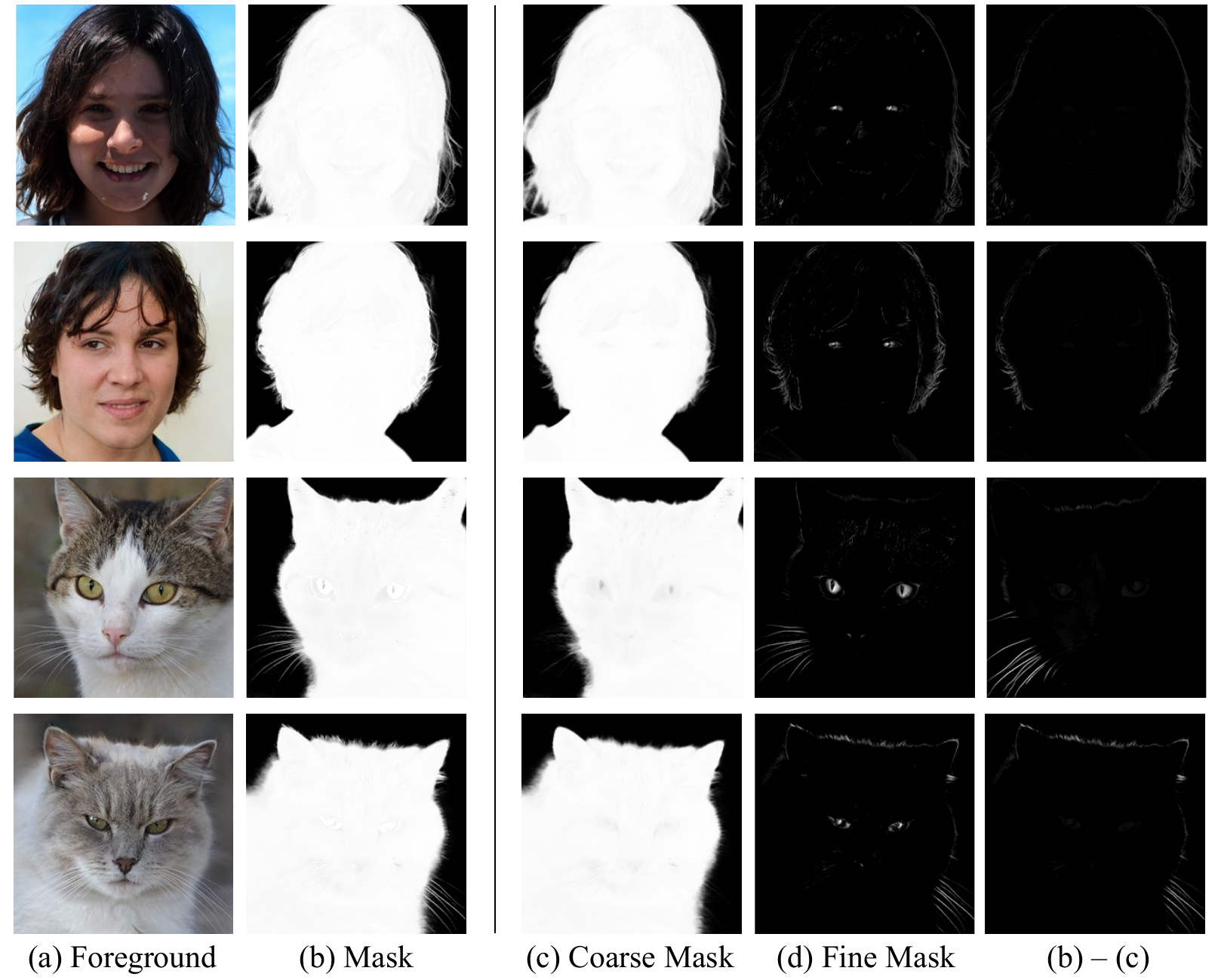}
    \caption{\textbf{Visualization of coarse and fine masks.} We generate a final mask by summing up coarse and fine masks and then clipping it to the range in [0,1]. Due to the clipping operation, the area where the fine mask contributes to the final mask is the difference between the final mask and the coarse mask.}
    \label{fig:masks}
    \vspace{50pt}
\end{center}
\end{figure}

We show some examples of coarse and fine masks in \fref{fig:masks}. As mentioned in \eref{eq:fineloss}, we penalize the area where the fine mask actually contributes to the final mask (the rightmost column of \fref{fig:masks}). Our generator can produce detailed alpha masks using the fine mask as needed. Finally, \fref{fig:gm_daux} illustrates architectures of the mask generator and the mask predictor. 

\section{Results on Unaligned Datasets}
\label{appendix:unaligned}

We also conducted training on unaligned datasets such as CUB and LSUN-Object. There are some changes in the training setting for this: 1) The coefficient of binarization loss is linearly reduced to 2.0 over the first 5K iterations. (default is 0.5). 2) We apply mask consistency loss after 5K iterations. 3) The average operation of the mask area loss is calculated for the mini-batch (not for each sample). 4) we set $\phi_1 = 0.2$ for LSUN-Object, and $\phi_1 = 0.1$ for CUB (\eref{eq:coarse}).

\begin{figure}[!hb]
\vspace{10pt}
\begin{center}
    \includegraphics[width=1\linewidth]{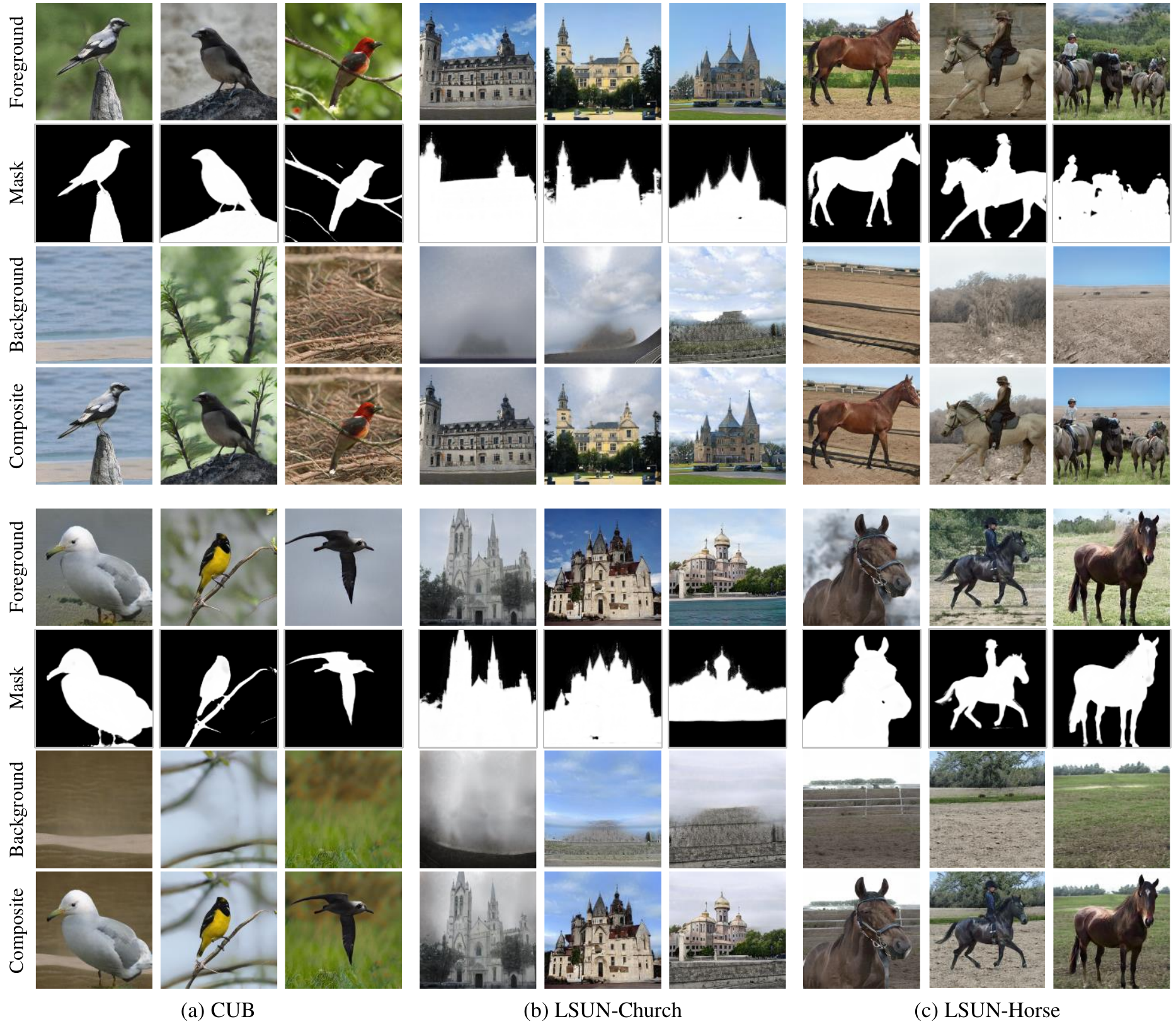}
    \caption{
    Curated qualitative results on unaligned datasets.
    }
    \label{fig:datasets}
\end{center}
\end{figure}

For LSUN-Object datasets, we use the first 100K images. We preprocess all datasets by center cropping and rescaling them to 256$\times$256. \fref{fig:datasets} shows the results of selected samples for three unstructured datasets.

\section{User Study on Mask Quality}
\label{appendix:user}

\begin{figure}[t]
\vspace{10pt}
\begin{center}
    \includegraphics[width=1\linewidth]{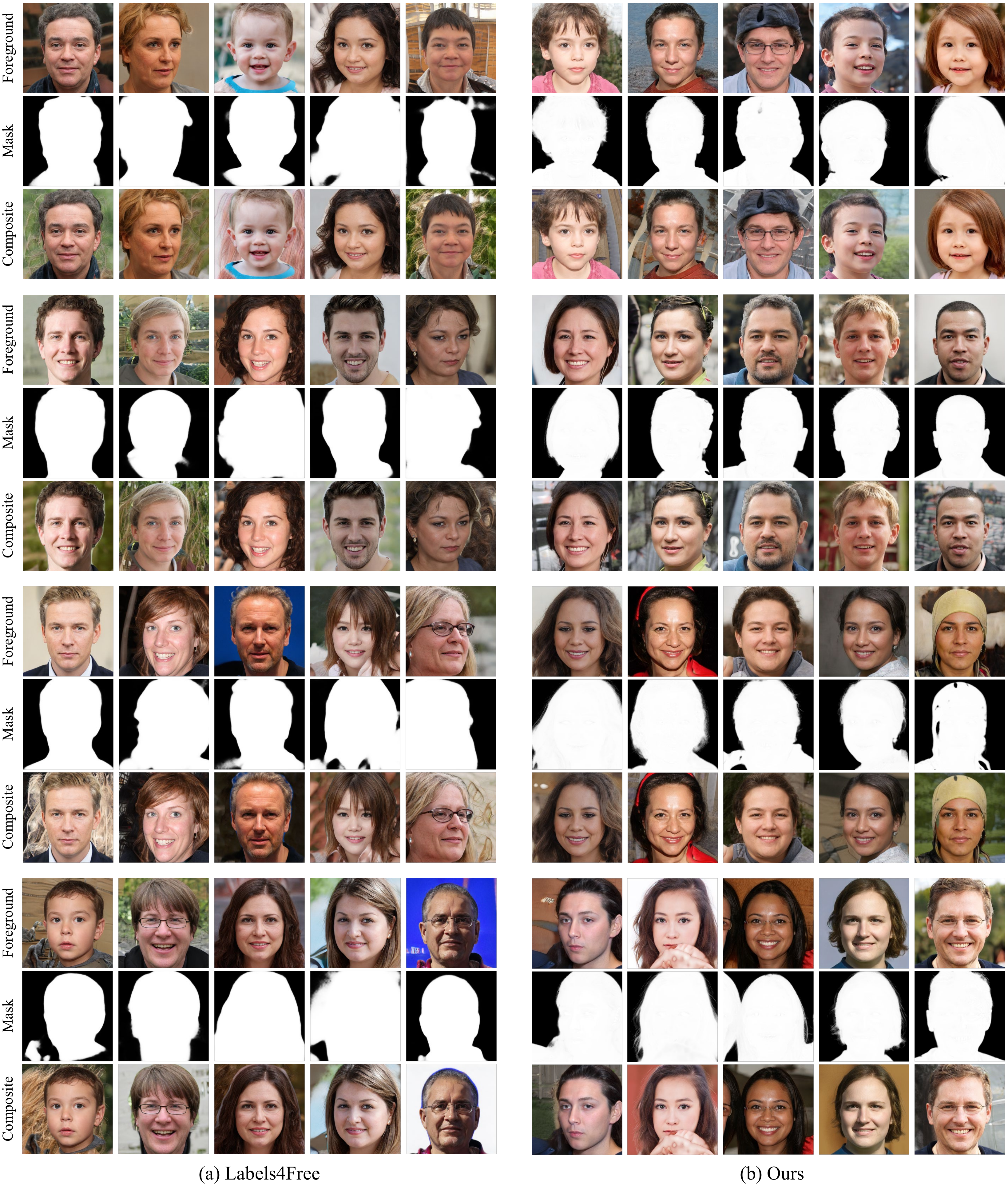}
    \caption{
    Uncurated qualitative comparison of image composition results on FFHQ, with truncation setting $\psi=0.7$.
    }
    \label{fig:uncurated_ffhq_07}
    \vspace{50pt}
\end{center}
\end{figure}

\begin{figure}[t]
\vspace{10pt}
\begin{center}
    \includegraphics[width=1\linewidth]{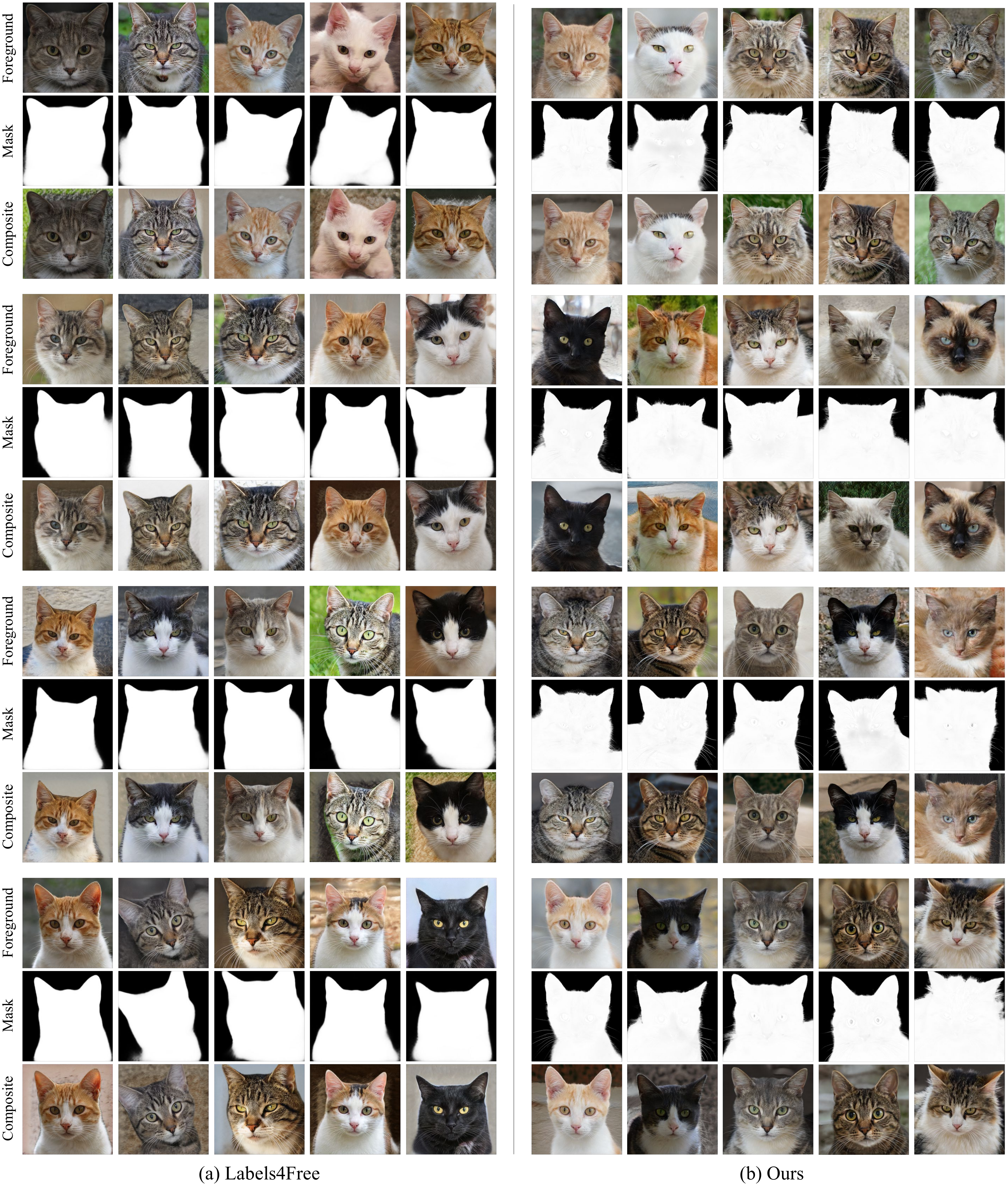}
    \caption{
    Uncurated qualitative comparison of image composition results on AFHQ, with truncation setting $\psi=0.7$.
    }
    \label{fig:uncurated_afhq_07}
    \vspace{50pt}
\end{center}
\end{figure}

To further evaluate the mask performance of our model, we asked 50 participants to choose more precise masks between ours and Labels4Free. 
In \tref{tab:userstudy} (a), we report the results for ten random matches of generated image-mask pair used in \fref{fig:uncurated_ffhq_07}-\ref{fig:uncurated_afhq_07}.
In \tref{tab:userstudy} (b), we report the results for the quality of masks obtained through the inversion of 20 real images (CelebAMask-HQ). For real image segmentation, both models were trained on FFHQ. Our model outperforms Labels4Free in mask quality of generated images and segmentation results of real images.

\begin{table}[!h]
\caption{ The reported values mean the preference rate of mask outputs from ours against Labels4Free. }
\centering
\begin{tabular}{c|cc|c}
 & \multicolumn{2}{c|}{(a) Generated} & \multicolumn{1}{c}{(b) Real} \\ \cline{2-4} 
 & $\quad$ AFHQv2-Cat & $\quad$ FFHQ $\quad$ & $\quad$ CelebA-HQ $\quad$ \\ \hline
Labels4Free & 15.8\% & 11.2\% & 11.8\% \\
Ours & 84.2\% & 88.8\% & 88.2\%

\end{tabular}

\label{tab:userstudy}
\end{table}

\end{appendix}
\end{document}